\documentclass[a4paper,fleqn]{cas-dc}

\usepackage[authoryear]{natbib}
\usepackage{url}
\usepackage{booktabs}
\usepackage{multirow}
\usepackage{algorithm}      
\usepackage{algpseudocode}  
\usepackage{algorithmicx}   
\usepackage{float}
\usepackage{multirow}       
\usepackage{xcolor}         
\usepackage{graphicx}       
\usepackage{booktabs}       
\usepackage{adjustbox}
\usepackage{tabularx}   
\usepackage{booktabs}   
\usepackage{soul}       
\usepackage{array}      
\newcolumntype{L}{>{\raggedright\arraybackslash}X}
\newcolumntype{C}{>{\centering\arraybackslash}X}
\usepackage{changepage}  

\newlength{\extralength}
\newlength{\fulllength}
\def\tsc#1{\csdef{#1}{\textsc{\lowercase{#1}}\xspace}}
\tsc{WGM}
\tsc{QE}

\begin{document}
\let\WriteBookmarks\relax
\def\floatpagepagefraction{1}
\def\textpagefraction{.001}

\shorttitle{}    

\shortauthors{}  

\title [mode = title]{GRAR: Glass-induced Reflection Artifact Removal in LiDAR Point Clouds}



\author[1]{Wanpeng Shao}[orcid=0009-0002-8395-8427]
\ead{wpshao@hnu.edu.cn}

\author[1]{Zeyi Guo}[orcid=0009-0006-5958-2422]
\ead{gzy@hnu.edu.cn}

\author[3]{Bo Zhang}[orcid=0009-0006-5958-2422]
\ead{zhangbo0720@hnu.edu.cn}

\author[2]{Yifei Xue}[orcid=0000-0002-4443-4367]
\ead{yifeixue@hnu.edu.cn}

\author[2]{Tie Ji}[orcid=0009-0005-0822-0600]
\ead{tieji@hnu.edu.cn}

\author[2]{Yizhen Lao}[orcid=0000-0002-6284-1724]
\cormark[1]

\affiliation[1]{organization={College of Computer Science and Electronic Engineering, Hunan University},
            city={Changsha},
            postcode={410012},
            country={China}}

\affiliation[2]{organization={School of Design, Hunan University},
            addressline={Pailou Road, South Campus, Taozihu},
            city={Changsha},
            postcode={410082},
            state={Hunan},
            country={China}}

\affiliation[3]{organization={School of Finance and Statistics, Hunan University, 109 Shijiachong Road, Financial \& Economic Campus, Yuelu District, Changsha, 410079, Hunan, China},
            city={Changsha},
            postcode={410012},
            country={China}}
            

 \cortext[1]{Corresponding authors: Yizhen Lao (yizhenlao@hnu.edu.cn)}

\begin{abstract}
Terrestrial Laser Scanning (TLS) point clouds captured in urban environments frequently suffer from glass-induced reflection artifacts, severely degrading downstream applications. Existing reflection artifact removal methods generally rely on ideal reflection symmetry assumptions, yet their performance is limited by inaccurate glass estimation and insufficient geometric representations. To address these issues, we propose a novel unified framework aimed at robust reflection artifact removal: In the first stage, we leverage a multi-modal vision foundation model to produce initial glass masks, which are then refined using geometric cues to achieve high-precision glass regions, followed by glass completion to recover missing regions caused by no-return measurements on transparent surfaces; In the second stage, we propose a physics-driven descriptor, termed Reflection-aware Local-Global Geometric Similarity (RE-LGGS), which is grounded in actual laser reflection geometry and jointly encodes multi-scale geometric structures and orientation consistency using PCA-based local shape representations, thereby significantly improving robustness against imperfect observations. Extensive experiments on multiple public TLS datasets demonstrate that our framework consistently outperforms state-of-the-art methods in reflection artifacts removal.
\end{abstract}

\begin{highlights}

\item Develops a vision foundation model-guided glass extraction and completion strategy that combines multimodal observations with geometric constraints to recover incomplete glass surfaces under severe measurement voids.

\item Develops the Reflection-aware Local-Global Geometric Similarity (RE-LGGS) descriptor, which jointly exploits multi-scale geometric structures and orientation consistency for robust virtual point detection under imperfect reflections.

\item Achieves better performance on multiple public TLS datasets, including dominant glass scenes, grid-like glass scenes, and subway scenes, demonstrating superior robustness against incomplete and distorted reflection observations.

\end{highlights}



\maketitle

\section{Introduction}
\label{sec:introduction}

Digital twins are the product of the deep integration of cutting-edge information technology with complex urban systems, increasingly becoming a core engine for advancing urban governance modernization and sustainable development. Their essence lies in constructing a digital mirror that synchronizes with the real city, enabling virtual-real interaction and intelligent intervention \citep{Ying2023New}.

Realizing this objective strictly demands large-scale, high-precision 3D geospatial data.
Currently, mainstream 3D data acquisition techniques can be broadly categorized into image-based reconstruction and LiDAR sensing. Although image-based~\citep{Snavely2006Photo,Furukawa2010Accurate} and neural rendering approaches~\citep{MildenhallNeRF2021,Kerbl20233D} have achieved impressive visual reconstruction quality, they often struggle to provide the metric-level geometric accuracy required by digital twin applications. In contrast, benefiting from its active sensing mechanism, LiDAR enables the direct acquisition of dense, centimeter-level 3D point clouds and remains the dominant technology for high-precision geospatial data collection in complex urban environments~\citep{Liu2023Literature}.

However, current LiDAR technology encounters severe limitations when operating in highly urbanized environments. To balance architectural esthetics with energy efficiency, modern buildings extensively utilize low-emissivity (low-emission) coated glass. While these advanced materials exhibit high transmittance in the visible spectrum, they demonstrate intense specular reflection characteristics within the near-infrared (NIR) band—the standard operating wavelengths for contemporary LiDAR sensors (e.g., 905~nm or 1550~nm). This discrepant optical behavior introduces severe signal degradation and reflection artifacts, thereby compromising the reliability of LiDAR perception systems that fundamentally rely on NIR backscatter detection~\citep{Fong2025Signs}.

\begin{figure}
    \centering
    \includegraphics[width=0.8\columnwidth]{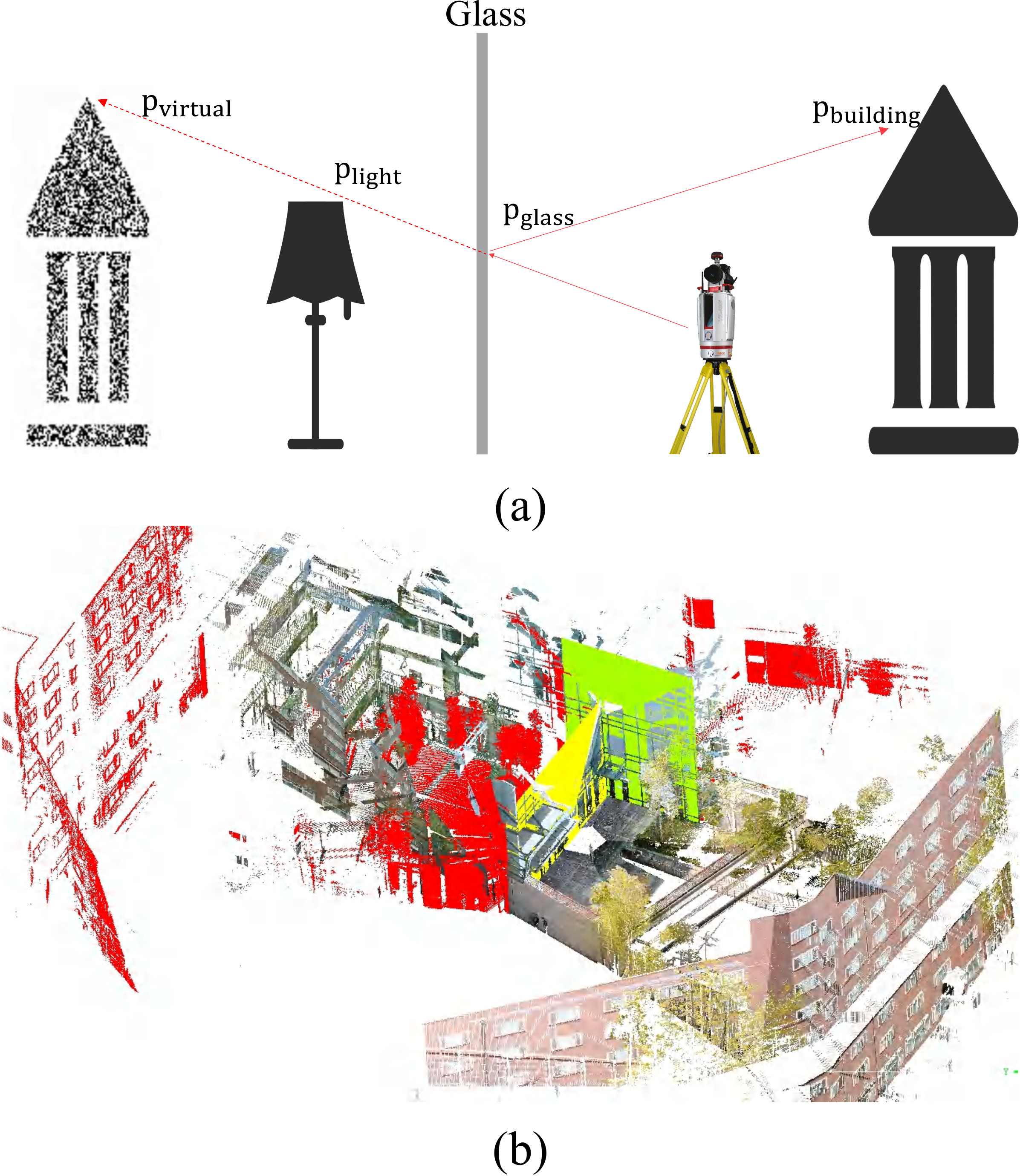}
    \caption{Reflection artifact in TLS point clouds. (a) The principle of reflection in TLS measurement. The laser beam hits the glass surface, producing a glass point ($P_{\mathrm{glass}}$), a virtual point reflected from a building by the glass ($P_{\mathrm{virtual}}$), and a light point inside the building ($P_{\mathrm{light}}$). (b) The real building scene captured by TLS where the glass planes are shown in yellow and green, and virtual points are shown in red.}
    \label{fig:graphical_abstract}
\end{figure}

The specific degradation mechanism is illustrated in Fig.~\ref{fig:graphical_abstract}: when a directional laser pulse strikes a glass surface (e.g., a curtain wall or storefront), it undergoes concurrent \textbf{transmission} and \textbf{reflection} at the optical interface. Specifically, a fraction of the laser beam penetrates the glass medium to hit the background structures, generating true return points from objects physically located behind the transparent barrier. Conversely, the specularly reflected laser beam is redirected into the foreground scene, subsequently projecting highly geometrically deceptive \textbf{reflection artifacts} (i.e.,\textbf{ virtual points}) that mirror real foreground objects onto the opposite side of the glass. These virtual points intertwine with valid measurements in 3D space, severely compromising the geometric consistency and fidelity of the captured geospatial data. Consequently, the resulting geometric corruption propagates to downstream tasks, including point cloud registration~\citep{dong2020registration}, 3D Building Information Modeling (BIM) reconstruction~\citep{Xiong2023Knowledge}, and semantic scene understanding~\citep{HAN2024500}.

Therefore, the reliable identification and elimination of glass-induced reflection artifacts are indispensable to ensuring the fidelity of 3D structural models for digital twin applications.

\subsection{Motivation}
\label{sec:Motivation}
While research on reflection removal in 3D scenarios remains relatively underexplored, existing point cloud denoising studies primarily focus on stochastic outliers that deviate from the true geometric surfaces of objects~\citep{Ambrosino2024Hybrid, GonizziBarsanti2024Evaluation, Zheng2022Single, Wang2024Point}. Such "geometric noise," typically originating from sensor measurement errors, environmental interference, or non-uniform sampling densities, is fundamentally distinct from glass-induced virtual artifacts. 

\begin{figure}
    \centering
    \includegraphics[width=0.98\columnwidth]{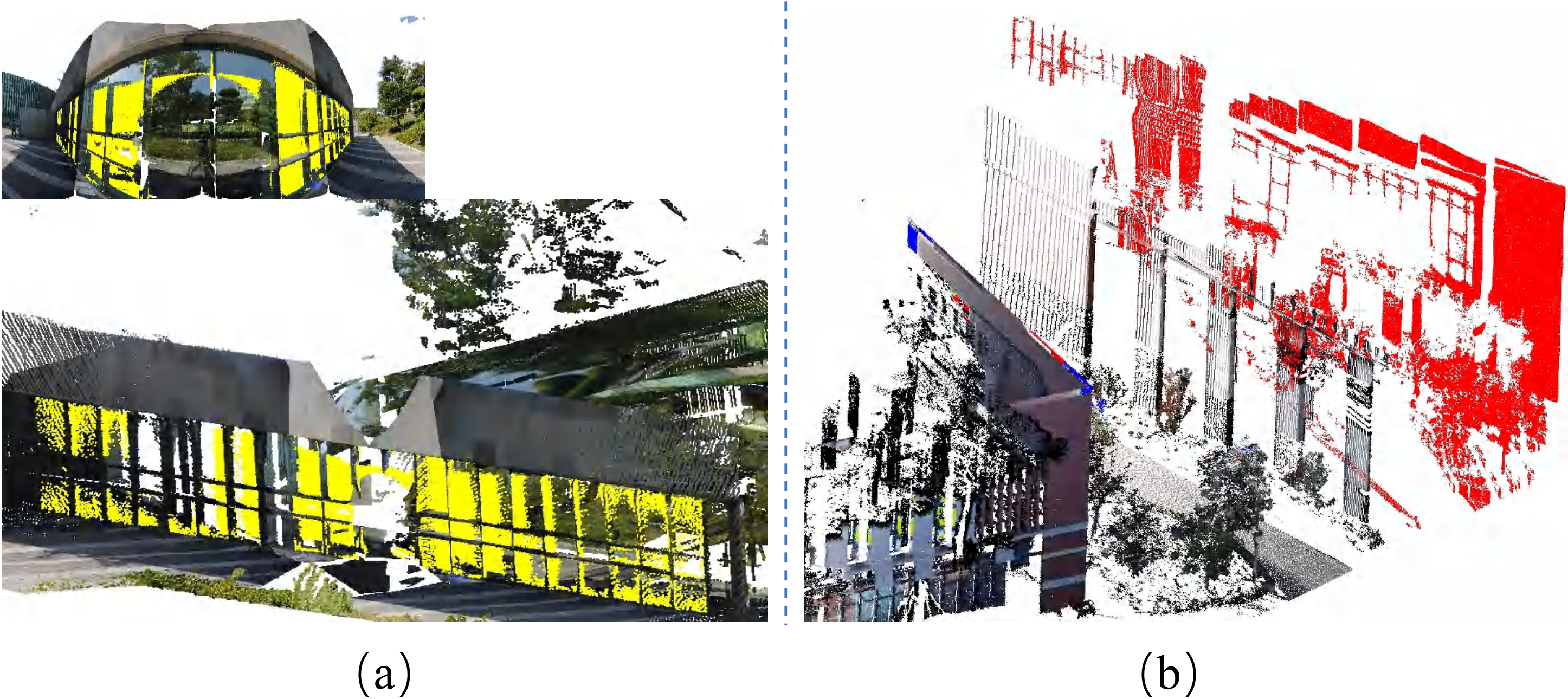}
    \caption{Challenges in reflection artifact removal (glass and virtual points are colored in yellow and red, respectively). (a) "Glass void" in TLS measurement. (b) TLS measurements with partial, distorted reflected virtual points.}
    \label{fig:challenges}
\end{figure}

Existing reflection removal methodologies~\citep{Yun2018Reflection,Yun2019Cluster, Yun2021Virtual, Shao2023Reflections, Fang2025Coupled, Shao2026GRASS} generally adhere to a two-stage framework: glass surface detection followed by virtual point identification via reflection symmetry and geometric similarity. Although they have reported promising results, their effectiveness largely depends on two implicit assumptions: (1) glass surfaces can be reliably detected from incomplete TLS observations, and (2) reflected observations preserve sufficiently accurate reflection symmetry with their real counterparts. Unfortunately, both assumptions are frequently violated in real-world urban environments, leading to substantial performance degradation.
Therefore, two critical physical challenges inherent to this paradigm remain inadequately addressed:

\textbf{1. The "Measurement Void" problem in TLS measurements.} 

Unlike conventional objects that produce dense and spatially continuous measurements, glass surfaces often generate sparse, fragmented, or even completely missing observations owing to their high transmissivity in Fig.~\ref{fig:challenges}(a). Consequently, existing methods typically treat glass detection as a purely geometric problem by exploiting indirect cues such as intensity and multi-echo responses. However, these geometric cues alone often provide insufficient evidence for reliable glass localization, making the recovery of complete glass structures from incomplete observations particularly challenging. As a result, virtual points produced from these undetected or fragmented glass regions cannot be accurately mapped, leading to incomplete virtual artifact removal. \textit{Therefore, the first challenge lies in how to detect and complete the glass structures from incomplete and sparse measurements, which serves as a prerequisite for effective virtual point removal.}

\textbf{2. The "Imperfect Reflection Symmetry" in virtual point identification:}

Existing virtual point identification methods generally assume strict reflection symmetry, where virtual points are expected to keep geometrically consistent mirrored structures with the real objects. However, real-world glass reflections rarely satisfy such ideal conditions. Due to inaccurate glass estimation, partial reflections, and internal multi-path reflections within the glass, the captured virtual points are often incomplete and geometrically distorted in Fig.~\ref{fig:challenges}(b). As a result, the geometric descriptors adopted by existing methods, which are inherently designed under ideal reflection symmetry assumptions, often experience severe performance degradation when confronted with incomplete and geometrically distorted reflection observations. \textit{Therefore, the second challenge lies in how to construct a robust geometric representation that can model imperfect reflection observations for reliable virtual point identification.}

\subsection{Contribution and Paper Organization}
\label{sec:Contribution_and_paper_organization}

To address the aforementioned challenges, we propose a novel unified framework, called GRAR in Fig.~\ref{fig:overall_workflow}, for robust glass-induced reflection artifact removal in TLS point clouds.

Regarding the first challenge, instead of treating glass detection as a purely geometric problem, we reformulate it as a cross-modal perception task. Specifically, panoramic RGB images captured by TLS-mounted, pre-calibrated camera together with TLS data-derived intensity and multi-count maps are jointly exploited within a vision foundation model to generate high precision glass masks. A subsequent glass extraction and completion strategy is further introduced to recover missing regions caused by no-return measurements on transparent surfaces.

Regarding the second challenge, to handle imperfect reflection observations, we propose a robust virtual point identification strategy called Reflection-aware Local-Global Geometric Similarity (RE-LGGS) descriptor. The proposed descriptor jointly models multi-scale geometric structures and orientation consistency using PCA-based representations, enabling reliable virtual point identification under incomplete and geometrically degraded reflections.

Our contributions can be summarized as follows:
\begin{itemize}
    \item We propose a cross-modal glass detection framework powered by a vision foundation model, which integrates panoramic RGB images with LiDAR points-derived modalities to achieve high-completeness glass extraction and glass region completion under no-return measurements.

    \item  We develop a Reflection-aware Local-Global Geometric Similarity (RE-LGGS) descriptor that jointly encodes multi-scale geometric structures and orientation consistency, enabling robust virtual point identification under imperfect reflection conditions.
    
    \item Extensive experiments on multiple public and TLS datasets demonstrate that the proposed framework consistently outperforms existing state-of-the-art methods in reflection artifact removal accuracy and point cloud quality preservation.
\end{itemize}

The rest paper is structured as follows. We first investigate existing denoising methods, including TLS-specific and indoor SLAM-related reflection artifacts removal in section~\ref{sec:related_work}.
Then, the design and implementation of the GRAR framework is introduced in section.~\ref{sec:method}. The efficacy of the proposed method is then rigorously validated through comprehensive experiments (Sec.~\ref{sec:Results}). Finally, we discuss the implications and limitations of our findings and present concluding remarks 
 (Sec.~\ref{sec:Discussion} and Sec.~\ref{sec:conclusion}).

\section{Related work}
\label{sec:related_work}

Current research on glass-induced reflection artifact removal can be broadly divided into studies conducted on static terrestrial laser scanning (TLS) data and those designed for mobile LiDAR systems. Owing to substantial differences in acquisition geometry, point density, and scene complexity, these two scenarios generally adopt different processing strategies.

\subsection{GRAR in TLS Measurements}
\label{sec:GRAS_TLS}

In TLS scenes, reflection artifacts often preserve strong geometric correlations with their corresponding real structures. Consequently, most existing approaches follow a two-stage workflow: (1) glass region estimation and (2) virtual point identification based on the estimated glass surfaces.

\subsubsection{Glass Region Estimation}

Existing glass detection methods can be broadly categorized into echo-based and intensity-based approaches.

\begin{itemize}
    \item \textbf{Echo-based methods}:When a laser beam interacts with a glass surface, multiple echoes may be generated from the glass interface, transmitted objects, and reflection-induced virtual points. Exploiting this phenomenon, several studies project TLS observations into image space and identify glass regions using echo-count information~\citep{Yun2018Reflection,Yun2021Virtual}. To improve spatial continuity, subsequent works further employ image segmentation or point cloud clustering techniques to aggregate sparse glass observations into larger surface regions~\citep{Yun2019Cluster,Lee2023Learning,Shao2026GRASS}. Nevertheless, because glass observations are frequently incomplete or entirely missing in real-world environments, accurately recovering continuous glass surfaces remains challenging.

    \item \textbf{Intensity-based methods}: Another line of research utilizes the relatively weak return intensity commonly observed on transparent glass surfaces~\citep{KASHANI201528099}. Based on this property, potential glass regions are identified through intensity-based screening and subsequent geometric verification~\citep{Fang2025Coupled}. However, the returned intensity is often influenced by background materials located behind the glass, particularly in the presence of highly reflective objects, reducing the stability of intensity-only cues~\citep{Shao2023Reflections}.

\end{itemize}

Overall, existing glass detection approaches primarily rely on incomplete LiDAR-derived observations, including echo and intensity cues, to infer glass locations. As a result, recovering complete glass surfaces under severe measurement voids remains a challenging problem.

\subsubsection{Virtual Point Identification}

After glass surfaces are estimated, virtual points are commonly detected through reflection symmetry and geometric similarity analysis across the glass plane. Existing methods can be roughly grouped into handcrafted feature-based approaches and learning-based approaches.

\begin{itemize}

\item \textbf{Handcrafted feature-based methods.}
Representative approaches employ local geometric descriptors such as FPFH~\citep{Yun2018Reflection,Yun2021Virtual,Shao2023Reflections,Yun2019Cluster} or local angular statistics~\citep{Fang2025Coupled} to measure the similarity between reflected observations and their corresponding real structures. These methods have demonstrated effectiveness in capturing local geometric characteristics. However, real reflection observations are frequently affected by sparsity, structural incompleteness, geometric distortion, and many-to-one correspondence effects, which substantially degrade the discriminative capability and reliability of these local descriptors.

\begin{figure*}
    \centering
    \includegraphics[width=\textwidth]{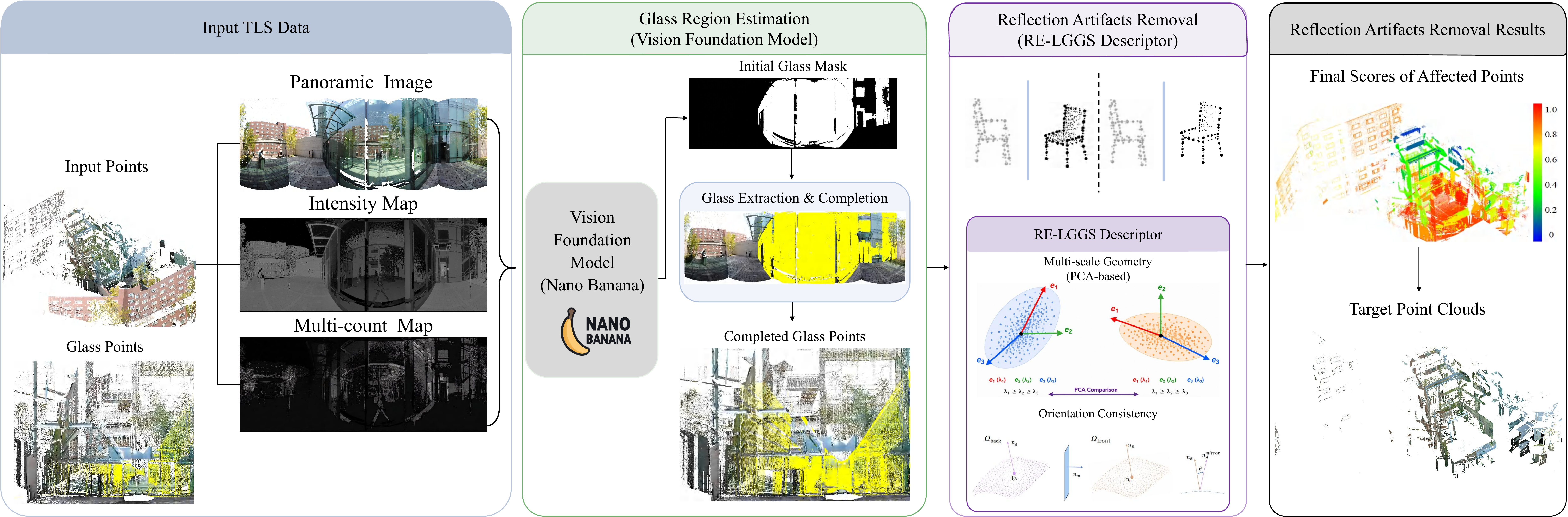}
    \caption{Glass-induced Reflection Artifact Removal in LiDAR Point Clouds (GRAR). Given input point clouds, we first project LiDAR data onto a spherical panoramic form. The RGB, intensity, and multi-count maps are then fed into a vision foundation model (e.g., NanoBanana) to generate accurate glass masks. These masks are back-projected into the original 3D space for extraction, refinement, and completion. Finally, the proposed RE-LGGS descriptor is used to identify reflection-affected points as virtual points.}
    \label{fig:overall_workflow}
\end{figure*}

\item \textbf{Learning-based methods.}
Recent studies formulate virtual point identification as a supervised learning problem using voxel-based networks~\citep{Lee2023Learning} or point-based architectures such as KPConv~\citep{KPConv2019,Shao2026GRASS}. These approaches are capable of learning complex geometric patterns from large-scale training data. However, constructing real TLS reflection datasets with reliable point-wise annotations is extremely expensive. Consequently, existing training datasets are typically generated using simplified reflection simulation strategies under idealized reflection assumptions, where virtual points are obtained through geometric reflection transformations followed by point sub-sampling. Such synthetic observations differ substantially from real reflection measurements, which potentially limit the generalization ability of learned models in practical environments.

\end{itemize}

Besides the conventional two-stage framework, several studies have explored multi-view consistency for reflection detection. For example, \citet{Gao2021Reflective,Gao2022Reflective} exploit the observation that real points remain geometrically consistent across multiple TLS viewpoints, whereas reflection artifacts vary with scanner positions. Although effective in controlled  indoor environments, these methods require multiple registered scans and are therefore less applicable to large-scale urban mapping scenarios.

\subsection{GRAR in Mobile LiDAR Systems}
\label{sec:Reflection_Removal_in_SLAM_scene}

Unlike TLS-based approaches, these methods are primarily designed for low-channel sensors operating in structured indoor environments, where reflections are often limited to mirrors or glass partitions and the scene scale remains relatively small.

\citet{Koch2016Detection,Koch2017} utilized multi-echo measurements to identify reflective surfaces through inter-echo distance analysis and subsequently extended the framework to 3D using mirror back-projection and ICP-based registration~\citep{Koch2017Detection}. Later, \citet{Zhao2020SSRR} incorporated dual-echo observations and intensity peak features to detect reflective surfaces and remove reflection artifacts through geometric consistency checks. Building upon this strategy, \citet{Li2024Detection} further extended the framework to complete indoor mapping scenes.

Although these methods have achieved promising performance in structured indoor environments, their assumptions regarding scene geometry, sensor configuration, and reflection behavior differ substantially from those encountered in large-scale TLS-based urban mapping applications.

\section{Methodology}
\label{sec:method}

\subsection{Overview}
\label{subsec:overview}

Fig.~\ref{fig:overall_workflow} shows the workflow of our approach, which mainly consists of two modules:

\begin{itemize}
    \item Glass regions estimation module (Section~\ref{subsec:glass_regions_estimation_module}): First, the TLS point cloud is projected onto a 2D scanning plane to generate a multi-count map and an intensity map, where each pixel records the echo count and the first-return intensity, respectively. Subsequently, intensity map, multi-count map together the synchronized panoramic RGB image, are jointly fed into a vision foundation model to extract a high-quality 2D semantic glass mask. Finally, the 2D glass mask is back-projected into the 3D space to isolate initial glass points, which are subsequently refined by geometric constraints and completed across measurement voids to fully restore continuous 3D glass surfaces.

    \item Reflection artifact removal module (Section~\ref{subsec:reflection_artifact_removal_module}): After glass detection, each point is symmetrically projected across the estimated glass plane to establish potential real-virtual correspondences. The Reflection-aware Local-Global Geometric Similarity (RE-LGGS) descriptor is then computed by jointly analyzing multi-scale geometric characteristics of virtual points, searched real points, and complete real structures. Finally, reflection symmetry and geometric similarity are combined to identify reflection-induced virtual points.
\end{itemize}

\begin{figure*}
    \centering
    \includegraphics[width=0.8\textwidth]{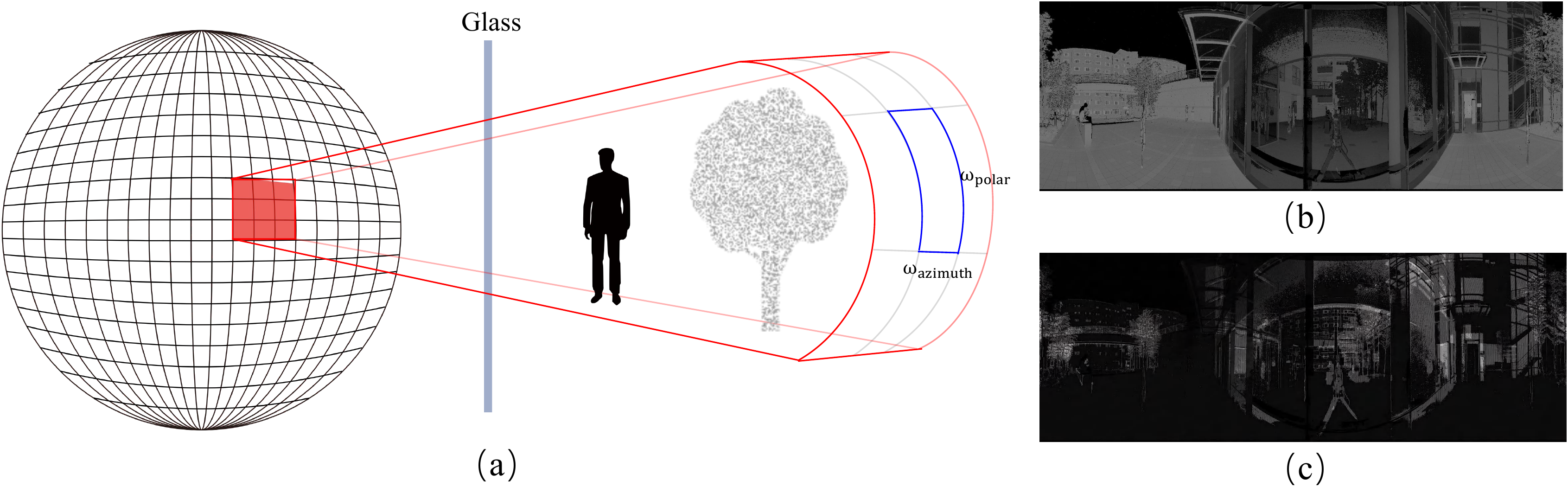}
    \caption{Spherical projection of TLS point clouds to produce intensity and multi-count map.}
    \label{fig:sphere_projection}
\end{figure*}
\begin{figure*}
    \centering
    \includegraphics[width=\textwidth]{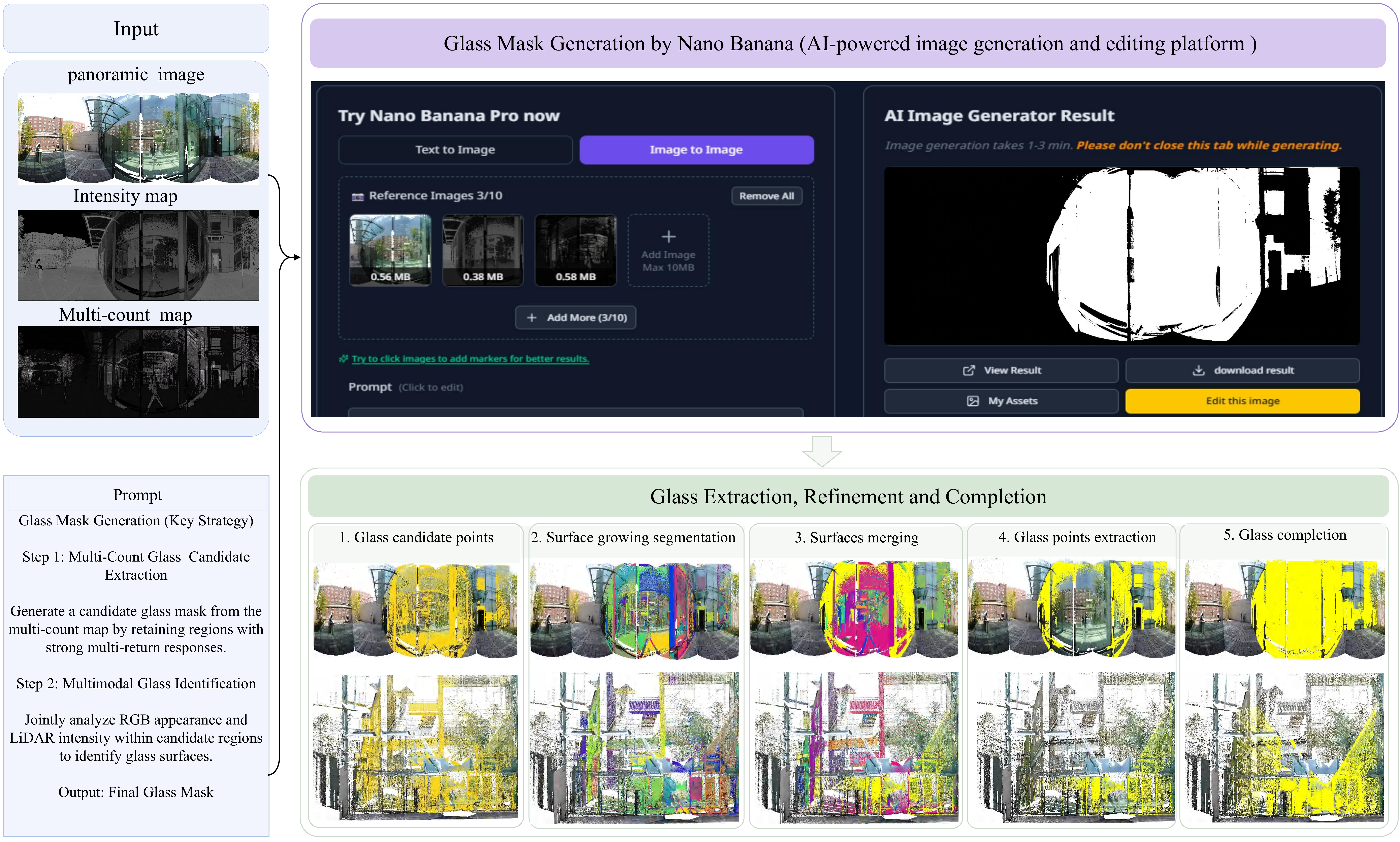}
    \caption{Overview of the proposed glass mask generation strategy. An initial glass mask is inferred from RGB imagery, LiDAR intensity maps, and multi-count maps using a vision foundation model, and is subsequently refined and completed through geometric constraints to obtain high-completeness glass surfaces.}
    \label{fig:glass_detection_module}
\end{figure*} 

\subsection{Glass Regions Estimation Module}
\label{subsec:glass_regions_estimation_module}

\subsubsection{Point Cloud Spherical Projection and Multi-Count Mapping}
\label{subsec:projection_and_mapping}

To facilitate cross-modal fusion with panoramic imagery, the raw TLS point cloud is transformed into a spherical image representation. As illustrated in Fig.~\ref{fig:sphere_projection}, a unit sphere centered at the scanner is discretized according to the scanner's intrinsic angular sampling intervals in the azimuth and polar directions. Instead of using individual angular bins, neighboring bins are aggregated into local $m \times m$ patches, each corresponding to a spatial frustum in the 3D scene. For every patch, the first-return intensity values and the total number of recorded echoes are accumulated to generate an intensity map and a multi-count map, respectively.

The patch size controls the trade-off between angular resolution and statistical robustness. Very small patches preserve fine spatial details but are sensitive to sampling fluctuations, whereas excessively large patches may blur structural boundaries. Through empirical evaluation, we found that $m=3$ provides a suitable balance. For a RIEGL VZ-400 scanner with an angular resolution of $0.06^{\circ}$, this configuration corresponds to an effective patch resolution of $0.18^{\circ}\times0.18^{\circ}$. The resulting spherical representation preserves large-scale scene structures while providing sufficiently stable measurements for subsequent glass-region estimation.

\subsubsection{Vision Foundation Model for Cross-Modal Glass Segmentation}
\label{subsec:Vision Foundation Model}

Prior to introducing vision foundation models, we briefly review existing specialized glass segmentation networks.

In the field of glass detection, research has evolved from traditional heuristic edge filters to end-to-end semantic segmentation networks, pioneered by GDNet \citep{Mei2020DontHitMe}, which treats glass as an independent semantic category. Subsequent studies have progressively enhanced recognition robustness by incorporating explicit edge context \citep{He2021Enhanced, FanWWWYM23}, specular reflection cues \citep{Lin2021Rich, Qi2024GlassMakesBlurs, Liu2024Multi}, and geometry-derived depth or semantic priors \citep{Lin2022Exploiting}. However, when confronted with static panoramic imagery in large-scale urban settings, these specialized networks exhibit severe vulnerability and frequently deliver unreliable or fragmented predictions due to two concurrent bottlenecks:

\begin{itemize}
    \item \textbf{Spatial-geometric domain gap.} Existing networks are inherently optimized for planar pinhole perspectives captured in constrained indoor scenes~\citep{Mei2020DontHitMe,Lin2021Rich,Lin2022Exploiting}. When applied to wide-field equirectangular projections ($360^{\circ}$ field-of-view), their limited receptive fields fail to capture long-range contextual dependencies across boundary disconnections, and their distortion-blind convolution kernels cannot adapt to latitude-dependent spherical distortions or extreme illumination fluctuations across wide panoramas~\citep{Zhang2024Behind, Tan2025DASCSPT}. 

   \item \textbf{Single-modality information bottleneck.} Glass-specific networks operating purely on RGB imagery are inherently challenged by the physical properties of glass surfaces. Transparent glass allows background structures to be directly observed through the surface, while specular reflections introduce visual patterns that closely resemble real objects, often leading to ambiguous semantic interpretations \citep{Lin2021Rich,Qi2024GlassMakesBlurs,Liu2024Multi}. Although recent methods have incorporated semantic priors or depth-related cues \citep{Lin2022Exploiting}, most existing approaches still rely predominantly on image appearance and therefore lack direct geometric observations of the scene. As a result, distinguishing glass from visually similar non-glass structures remains difficult in large-scale urban environments.

\end{itemize}

As a result, distinguishing glass from visually similar non-glass structures remains challenging, highlighting the need for multi-modal approaches.

The paradigm shift to multi-modal Vision Foundation Models (VFMs), specifically the lightweight \textbf{Nano-Banana} framework, directly addresses these limitations through joint semantic reasoning across heterogeneous visual modalities. The effectiveness of the proposed \textbf{Nano-Banana}-driven module stems from two fundamental capabilities.

First, \textbf{Nano-Banana} leverages high-level semantic priors acquired through large-scale cross-modal pre-training~\citep{zuo2025nanobananaprolowlevel}. Recent studies have shown that foundation models encode not only semantic concepts, but also rich structural information, including object boundaries, scene layouts, and contextual relationships~\citep{rombach2022high}. Consequently, when processing panoramic RGB imagery together with LiDAR-derived intensity and multi-count maps, the model can reason about glass as a coherent scene component rather than relying solely on local appearance cues. This capability is particularly beneficial in regions where glass observations are sparse, incomplete, or partially obscured.

Second, \textbf{Nano-Banana} provides strong text-guided semantic decomposition capabilities~\citep{Qian2025Pico}. Glass detection in urban environments is inherently challenging because transparent surfaces, reflections, and background structures are visually intertwined. Through prompt-conditioned reasoning and cross-attention mechanisms, the model can separate glass regions from surrounding scene content while simultaneously exploiting complementary information provided by the intensity and multi-count maps~\citep{zhang2023adding}. As a result, the generated masks exhibit improved completeness and boundary consistency compared with conventional RGB-based glass segmentation approaches.

From a practical geometric engineering perspective, the proposed framework offers two additional advantages:

First, it eliminates the need for collecting and annotating large-scale TLS-specific glass datasets. Acquiring reliable ground-truth labels for glass surfaces and reflection artefacts across diverse urban environments is both labor-intensive and economically impractical. By exploiting the zero-shot generalization capability of foundation models~\citep{zuo2025nanobananaprolowlevel}, the proposed approach can be directly applied to previously unseen scenes without task-specific retraining or parameter tuning.

Second, the proposed framework separates semantic glass identification from geometric reconstruction. The role of \textbf{Nano-Banana} is to provide a high-completeness glass mask in the image domain by jointly reasoning over panoramic RGB imagery, LiDAR intensity maps, and multi-count maps, as illustrated in Fig.~\ref{fig:glass_detection_module}. The generated mask subsequently serves as the basis for the geometric refinement and completion procedures in 3D space. This design allows the framework to benefit from the semantic reasoning capability of foundation models while preserving the metric and structural consistency required for TLS-based surveying and mapping applications.

Therefore, rather than developing another task-specific glass segmentation network, we exploit a vision foundation model as a universal semantic prior to bridge the observation gap caused by incomplete TLS measurements.

\subsubsection{Glass Extraction, Refinement and Completion}
\label{subsec:glass_detection_Refinement}

Upon acquiring the 2D glass mask, the pixels within the masked regions are back-projected into the 3D spatial domain as candidate glass points. Due to missing glass regions, even real objects within the projected 3D frustum inside the building may be misidentified as candidate glass points, as illustrated in step 1 of Fig.~\ref{fig:glass_detection_module}. Under such conditions, iterative RANSAC is unreliable because the high proportion of non-glass points dominates the plane fitting~\citep{xu2019pairwise}.

We adopt a classical surface growing segmentation method~\citep{Vosselman2013ISPRS} to partition candidate glass points into planar structures. Segmentation is initialized from planar seeds detected via a 3D Hough transform~\citep{Vosselman2004Recognising} and subsequently expanded through iterative surface growing. Points belonging to the same planar region are assigned a unique segment ID, as shown in step 2 of Fig.~\ref{fig:glass_detection_module}. To filter non-glass artifacts, adjacent coplanar clusters are merged based on normal vector consistency and point-to-plane distance thresholds, forming dominant planar segments (step 3 of the Fig.~\ref{fig:glass_detection_module}).

Invalid segments are then discarded through a two-fold geometric validation:

\begin{enumerate}
    \item \textbf{Structural scale constraint:} segments with low point counts, high local curvature, or high linearity—typical of vegetation or noise—are removed.
    
    \item \textbf{Half-space constraint:} let the candidate glass plane be $\Pi: ax+by+cz+d=0$ and the scanner position $o$. For each point $p$, define
    \begin{equation}
        f(x) = a x + b y + c z + d.
    \end{equation}
    Points for which $f(o) f(p) < 0$ lie on the opposite side of the scanner relative to the plane and are discarded.
     After iterating over all candidates, only the  glass points remains in step 4 of Fig.~\ref{fig:glass_detection_module}.
\end{enumerate}

Finally, we perform a geometry-based completion for points that are recognized as belonging to glass regions but are sparsely sampled.

Specifically, for each point $p_m$ that falls within the estimated glass mask, we cast a ray from the scanner origin $\mathbf{o}$ through $p_m$ and compute its intersection with the glass plane $\Pi$. The intersection parameter $t^*$ is obtained by solving $\mathbf{n}\cdot(\mathbf{o} + t(p_m-\mathbf{o})) + d = 0$:
\begin{equation}
t^* = -\frac{\mathbf{n}\cdot\mathbf{o} + d}{\mathbf{n}\cdot(p_m - \mathbf{o})}.
\end{equation}
The completed glass point is then
\begin{equation} 
p_g = \mathbf{o} + t^* (p_m - \mathbf{o}).
\end{equation}
Only candidate points located inside the projected glass mask are completed, preserving measurement fidelity. The final 3D glass reconstruction is shown in Step 5 of Fig.~\ref{fig:glass_detection_module}.

\begin{figure}
    \centering
    \includegraphics[width=0.98\columnwidth]{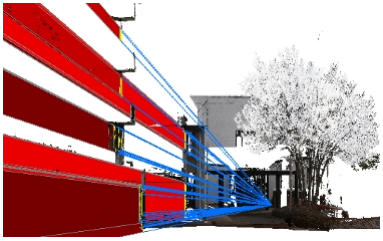}
    \caption{Affected area detection in real scene. Line in blue color is the transmission path from the scan pose to the outline of each glass region. Point fall into the red area are reflection-affected points.}
    \label{fig:affected_area_detection}
\end{figure}

Once glass surfaces are established, the corresponding reflection-affected point clouds can be extracted by tracing the laser transmission path from the scanner through the four corners of each glass plane in Fig.~\ref{fig:affected_area_detection}.

\subsection{Reflection artifact removal module}
\label{subsec:reflection_artifact_removal_module}

In this section, we focus on detecting virtual points in affected areas estimated in
section~\ref{subsec:glass_regions_estimation_module}.Based on the geometric characteristics of the TLS data, we choose to combine reflection symmetry and geometric similarity to detect virtual reflection points. 

\subsubsection{Reflection symmetry}
\label{subsubsec:Reflection_symmetry}
For a given point $p \in \Omega_{\mathrm{back}}$, we compute a symmetry score $\gamma_{\mathrm{symmetry}}(p)$ that reflects the proximity between the symmetric position $\hat{p}$ of $p$ and its nearest counterpart in $\Omega_{\mathrm{front}}$. The reflected point $\hat{p}$ across the glass plane $\Pi$ is obtained via a Householder transformation~\citep{Householder1958}. The Householder matrix $\mathbf{A}$ is defined as $\mathbf{A} = \mathbf{I} - 2\mathbf{n}\mathbf{n}^\top$, where $\mathbf{I}$ is the identity matrix and $\mathbf{n}$ is the unit normal vector of the plane. For a plane $ax + by + cz + d = 0$, the transformation matrix in homogeneous coordinates is:
\begin{equation}
\mathbf{A} = 
\begin{bmatrix}
    1 - 2a^2 & -2ab     & -2ac     & -2ad \\
    -2ab     & 1 - 2b^2 & -2bc     & -2bd \\
    -2ac     & -2bc     & 1 - 2c^2 & -2cd \\
    0        & 0        & 0        & 1
\end{bmatrix}.
\end{equation}
The Householder matrix is orthogonal. The symmetric counterpart of $p$ is then $\hat{p} = \mathbf{A}_{\Pi} p$, where $\mathbf{A}_{\Pi}$ corresponds to $\Pi$, with $p$ and $\hat{p}$ in homogeneous coordinates. We then use a $k$-d tree to find the nearest point $q \in \Omega_{\mathrm{front}}$ to $\hat{p}$. The symmetry score is defined as an exponential decay of this distance:
\begin{equation}\label{eq:symmetry}
    \gamma_{\mathrm{symmetry}}(p) = \exp \left( -\frac{\|\hat{p} - q\|}{\beta_1} \right),
\end{equation}
where $\beta_1$ controls the sensitivity to positional deviations. A larger $\beta_1$  reduces the sensitivity of the symmetry score to positional deviations, making the scoring function more tolerant to misalignments.

\begin{figure}
    \centering
    \includegraphics[width=0.98\columnwidth]{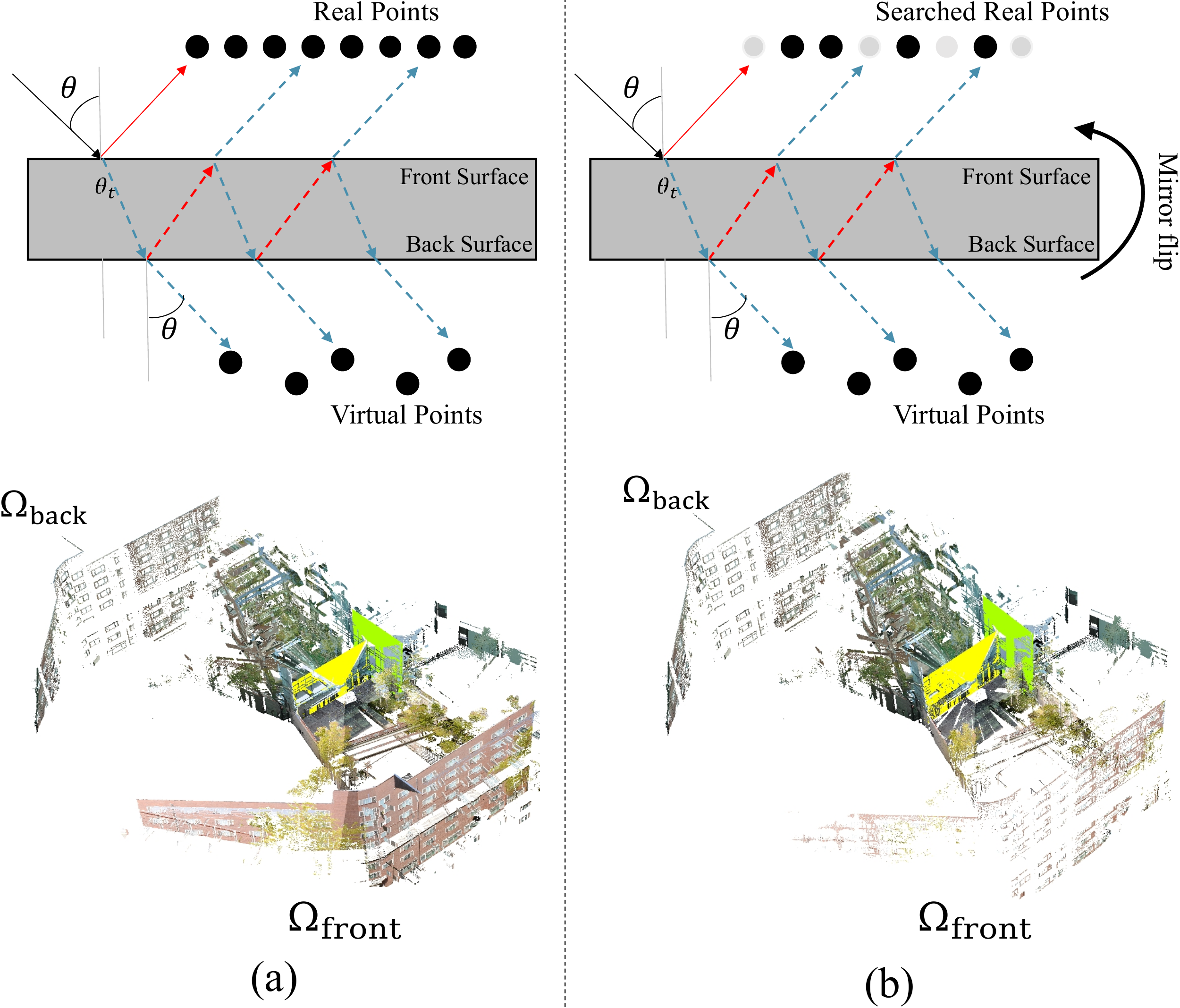}
    \caption{Depiction of internal multiple reflections and their impacts. (a) Illustration of multiple reflections inside glass structures and the resulting TLS observations, where reflection-induced virtual points become sparse, geometrically distorted, and spatially misaligned. (b) Searched real points retrieved through symmetric re-projection of the virtual points followed by nearest-neighbor search. Owing to reflection imperfections and geometric inconsistencies, many virtual points (gray points in the upper panel) fail to establish valid correspondences. Consequently, the retrieved real points in front of the glass are significantly sparser and more incomplete than the virtual observations.}
    \label{fig:reflection_artifact_removal_module_1}
\end{figure}

\subsubsection{Geometric similarity}
\label{subsubsec:Geometric similarityr}

Under ideal conditions, a reflection artifact should perfectly mirror a real structure across the glass plane. However, as illustrated in Fig.~\ref{fig:reflection_artifact_removal_module_1}(a), real-world glass reflections rarely satisfy this ideal symmetry due to complex physical anomalies such as internal multi-path reflections. Consequently, the reflected observations typically exhibit two major degradation characteristics:
(i) \textbf{Geometric incompleteness}, characterized by severe sparsity, structural fragmentation, and substantial loss of fine geometric details;
(ii) \textbf{Spatial distortion}, manifested as geometric inconsistencies with their corresponding real structures even after specular projection.
Moreover, as illustrated in Fig.~\ref{fig:reflection_artifact_removal_module_1}(b), such imperfect reflections inevitably introduce a many-to-one mapping anomaly, where multiple distorted virtual points may correspond to the same real-world point.

To address this issue, we first introduce the concept of searched real points. For each virtual point $p$, its mirrored position $\hat{p}$ is projected across the estimated glass plane, and the nearest neighbor in the complete real point cloud is selected as the corresponding searched real point $q$.

Unlike complete real structures, searched real points naturally inherit the many-to-one correspondence pattern caused by imperfect reflections. Consequently, they exhibit sparsity and structural degradation characteristics that are more consistent with the virtual observations, providing a geometrically balanced reference for comparison. Nevertheless, this degradation also leads to the loss of fine-scale geometric details.

\begin{figure}
    \centering
    \includegraphics[width=0.98\columnwidth]{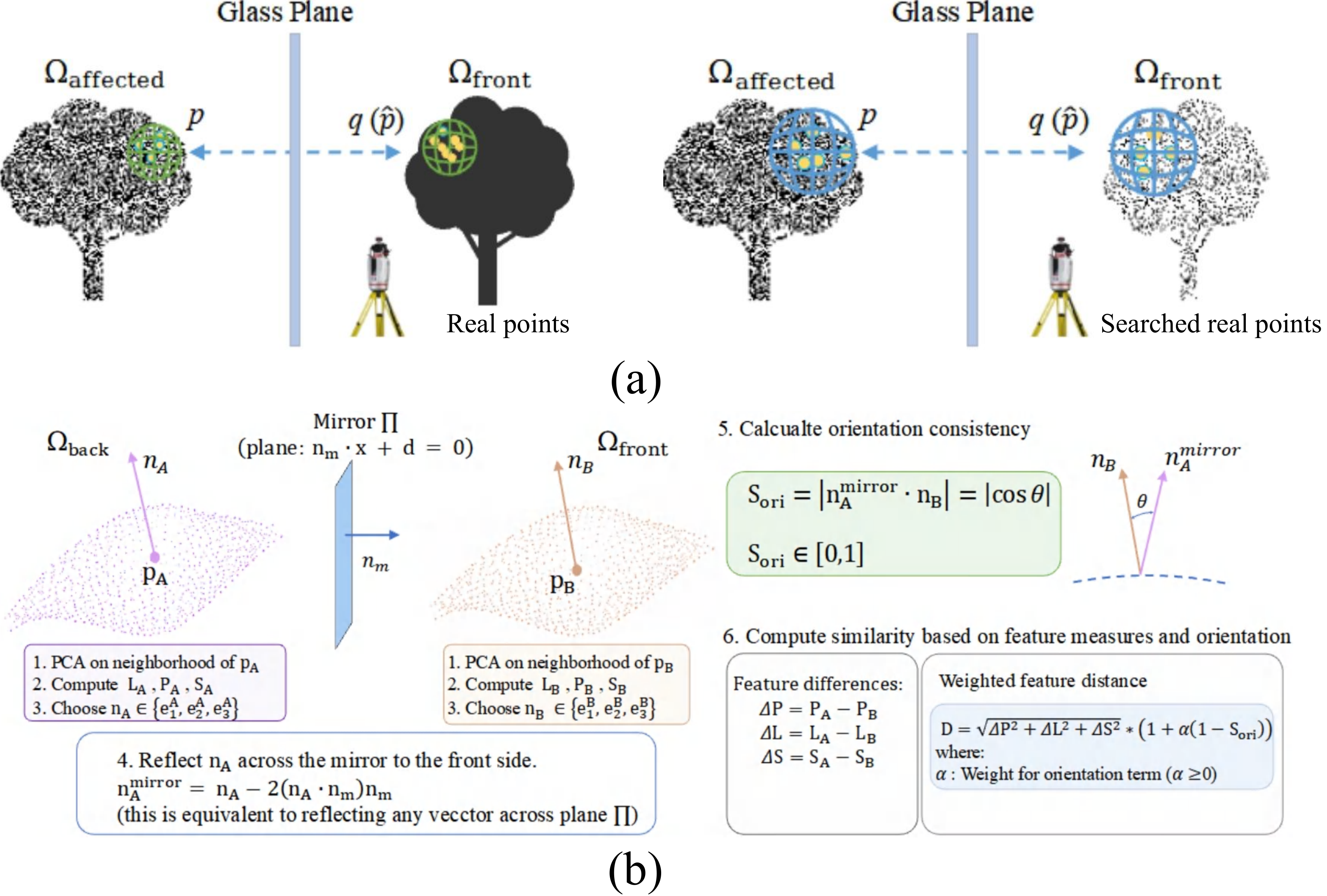}
    \caption{Overview of the proposed RE-LGGS descriptor for imperfect reflection observations. (a) Visualization of reflection observations between direct real-point correspondence and the searched real points. (b) Illustration of the PCA-based geometric descriptor with orientation consistency constraints at a single neighborhood scale. The descriptor is computed in the same manner at multiple scales.}
    \label{fig:reflection_artifact_removal_module_2}
\end{figure}

To simultaneously preserve local geometric fidelity and global structural consistency, the proposed RE-LGGS descriptor adopts a dual-scale collaborative strategy with two support radii, namely a local radius $r_1$ and a global radius $r_2$:
\begin{itemize}
    \item \textbf{Small radius neighborhoods} extracted from the complete local real points to preserve high-fidelity local geometric details (Fig.~\ref{fig:reflection_artifact_removal_module_2}(a));
    \item \textbf{Large radius neighborhoods} extracted from the 
    searched real points to model robust global structural skeletons under severe sparsity (Fig.~\ref{fig:reflection_artifact_removal_module_2}(b)).
\end{itemize}
Note that for the virtual points, both the small- and large-radius features are computed directly from its own local neighborhoods.

Instead of pursuing highly discriminative but potentially unstable high-dimensional descriptors, this work prioritizes PCA-based low-order geometric statistics. Unlike histogram-based descriptors such as FPFH, PCA-derived geometric attributes remain relatively stable under severe point sparsity, structural incompleteness, and geometric distortions, which are common characteristics of reflection-induced virtual observations. This robustness makes PCA particularly suitable for modeling imperfect reflections encountered in real TLS environments, as illustrated in Fig.~\ref{fig:reflection_artifact_removal_module_2}(b)
Let $\mathcal{N}(p)$ be the $r$-radius neighbor set of point $p$. Perform PCA on the covariance matrix
\begin{align}
    \mathbf{C} = \frac{1}{k} \sum_{q \in \mathcal{N}} (q - \bar{q})(q - \bar{q})^\top
\end{align}
yielding eigenvalues $\lambda_1 \ge \lambda_2 \ge \lambda_3 \ge 0$. We define the structural attributes as:
\begin{equation}
    L = \frac{\lambda_1 - \lambda_2}{\lambda_1}, \quad
    P = \frac{\lambda_2 - \lambda_3}{\lambda_1}, \quad
    C = \frac{\lambda_3}{\lambda_1},
\end{equation}
where $L$, $P$, and $C$ denote linearity, planarity, and curvature (scattering), respectively. By extracting these attributes across the dual neighborhood scales, the geometric profile for each point is formalized as:
\begin{equation}
    \mathbf{D} = [L_1, P_1, C_1, \; L_2, P_2, C_2],
\end{equation}
where subscripts $1$ and $2$ indicate the small- and large-radius scales, respectively.

Treating these attributes as coordinates in a low-dimensional shape space, the base geometric discrepancies between two descriptors $\mathbf{D}_a$ and $\mathbf{D}_b$ are evaluated separately for each scale via Euclidean distance. The Euclidean distance is adopted because the PCA-derived attributes are normalized and reside in a compact low-dimensional shape space. Under this representation, Euclidean distance provides a stable, computationally efficient, and physically interpretable measure of geometric discrepancy while avoiding the sensitivity and parameter dependence often associated with high-dimensional histogram matching:
\begin{align}
    D_{g1} &= \sqrt{(L_1^a-L_1^b)^2 + (P_1^a-P_1^b)^2 + (C_1^a-C_1^b)^2}, \\
    D_{g2} &= \sqrt{(L_2^a-L_2^b)^2 + (P_2^a-P_2^b)^2 + (C_2^a-C_2^b)^2}.
\end{align}

Nevertheless, relying solely on shape geometry is still insufficient when encountering distinct semantic structures with similar low-order statistics, such as orthogonal vertical walls and horizontal floors. To robustly resolve this ambiguity, we incorporate an adaptive orientation consistency constraint. Rather than enforcing a uniform directional alignment, the dominant orientation vector $\mathbf{e}$ of a neighborhood is adaptively selected based on its primary geometric tendency:
\begin{equation}
\mathbf{e} =
\begin{cases}
    \mathbf{e}_1 & \text{if } L \ge P \text{ and } L \ge C \quad (\text{linear}), \\
    \mathbf{e}_3 & \text{if } P \ge L \text{ and } P \ge C \quad (\text{planar}), \\
    \mathbf{0}   & \text{otherwise (isotropic)}.
\end{cases}
\end{equation}
Here, $\mathbf{e}_1$ and $\mathbf{e}_3$ represent the principal eigenvector and the surface normal vector obtained from PCA, respectively. This conditional logic ensures that meaningful directional constraints are preserved for linear and planar geometries, while unreliable orientations are strictly deactivated ($\mathbf{0}$) for isotropic environments.

The spatial orientation consistency between two adaptively selected orientation vectors $\mathbf{e}_a$ and $\mathbf{e}_b$ is measured by their absolute dot product:
\begin{equation}
    S_{\text{ori}} = |\mathbf{e}_a^\top \mathbf{e}_b|.
\end{equation}

The final orientation-aware geometric discrepancy
$D_{\text{RE-LGGS}}$ is then formulated as:
\begin{equation}
    D_{\text{RE-LGGS}} = D_{g1}\bigl(1+\alpha(1-S_{\text{ori},1})\bigr) + D_{g2}\bigl(1+\alpha(1-S_{\text{ori},2})\bigr),
\end{equation}
Where $\alpha$ is a weighting parameter governing the sensitivity to orientation misalignment. In our implementation, the penalty factor $\alpha$ is used to balance the significance of geometric shape and spatial orientation. Under this formulation, larger spatial angular differences yield severe numerical penalties, effectively suppressing false correspondences between geometrically similar but directionally inconsistent structures.

Ultimately, the reflection-aware structural similarity is mapped via an exponential kernel:
\begin{equation}\label{eq:similarity}
    \gamma_{\mathrm{similarity}}(p) = \exp\left(-\frac{D_{\text{RE-LGGS}}}{\beta_2}\right),
\end{equation}
where $\beta_2$ controls the sensitivity of the geometric similarity. A larger $\beta_2$ yields a smoother mapping, making the geometric similarity less sensitive to variations in $D_{\text{RE-LGGS}}$, while a smaller $\beta_2$ imposes a stricter geometric consistency constraint.

Compared with conventional local descriptors, the proposed RE-LGGS descriptor prioritizes low-order structural robustness over high-dimensional discriminateness. By seamlessly integrating dual-scale geometric statistics with adaptive orientation penalties, it provides an exceptionally stable and physically interpretable metric for robust virtual point identification under severe glass reflection degradations.

\subsubsection{Virtual points detection}
\label{subsubsec:Virtual_points_detection}

\begin{figure*}
    \centering
    \includegraphics[width=\textwidth]{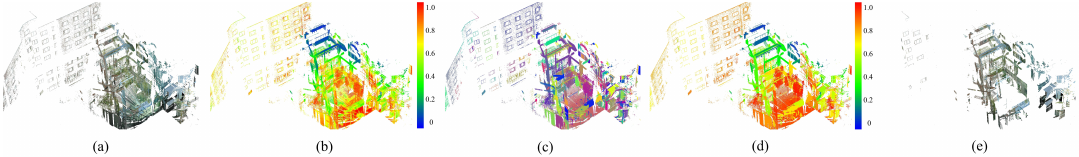}
    \caption{Overview of virtual points removal process. (a) Reflection-affected points. (b) Point-level final scores of affected points where points differ in the  same shape. (c) Segmentation results. (d) Segment-level final scores of affected points where points in the same  shape show the same score. (e) Reflections removal results.}
    \label{fig:reflections_removal_process}
\end{figure*}

We combine the symmetry score in Eq.~\eqref{eq:symmetry} and the similarity score in Eq.~\eqref{eq:similarity} to compute a final point-wise score $\gamma(p)$. A multiplicative formulation is adopted because reflection symmetry and geometric similarity represent two complementary necessary conditions for virtual point identification. A point exhibiting strong symmetry but poor geometric consistency, or vice versa, should not be regarded as a reliable virtual point. The product formulation naturally suppresses such ambiguous cases while emphasizing points that simultaneously satisfy both criteria:
\begin{equation}
    \gamma(p) = \gamma_{\text{symmetry}}(p) \cdot \gamma_{\text{similarity}}(p),
\end{equation}

By adjusting the parameters $\beta_1$ and $\beta_2$ in Eqs.~\eqref{eq:symmetry} and~~\eqref{eq:similarity}, respectively, the final scores of affected points can be reasonably distributed.

Fig.~\ref{fig:reflections_removal_process}(b) shows the point-level final scores of affected points, where we find that points belonging to the same physical structure often receive varying scores.
In such a case, it is challenging to select an appropriate score threshold to filter the reflection artefacts in a complete from the TLS data.
To overcome this issue, we formulate an energy function given by:
\begin{equation}
\arg \min_{g \in \mathcal{G}} \sum_i \| v_i - f_i \|^2 + \mu \sum_{(i,j) \in E} w_{ij} [g_i - g_j \neq 0]
\label{eq:optimization}
\end{equation}
Where $V$ represents the input set of $N$ 3D points, and each point is associated with a feature vector
$f_i=[L_i,P_i,C_i]$ composed of three PCA-derived dimensionality attributes, namely linearity, planarity, and scattering.
These features have been widely adopted as effective geometric constraints for point cloud segmentation and urban scene classification due to their robustness and discriminative capability~\citep{Landrieu2017}.
The first term describes the fidelity of resulting segments of corresponding local features $f_i$. The second term $w_{ij} [.]$ is the Iverson bracket, which determines if the neighboring components are merged or not. The factor $\mu$ measures the smoothness of neighboring resulting partitions. The problem can be solved with a few graph-cut iterations~\citep{Landrieu2017CutPursuit}.

After segmentation in Fig.~\ref{fig:reflections_removal_process}(c), we compute the average final score for each segment. Consequently, all points within the same segment share an identical score, as illustrated in Fig.~\ref{fig:reflections_removal_process}(d).

Finally, we fine-tune the final score threshold $\bar{\gamma}$ to separate virtual points from real points by assigning a binary label $l_i$ to each point $p_i$:
\begin{equation}
    l_i = 
    \begin{cases} 
        1, & \text{if } \gamma(p_i) > \bar{\gamma} \quad (\text{virtual point}), \\
        0, & \text{if } \gamma(p_i) \leq \bar{\gamma} \quad (\text{real point}).
    \end{cases}
\end{equation}

Such that $ l_i = 1 $ when $ p_i $ is virtual and $ l_i = 0 $ when $ p_i $ is real. The reflections removal result is shown in Fig.~\ref{fig:reflections_removal_process}(e).

\section{Experiment}
\label{sec:Results}

\subsection{Experimental Setup}
\label{sec:experimental_setup}

To evaluate our method’s performance on real-world data in diverse reflection scenarios, we conducted experiments using three distinct TLS datasets: UNIST
Building Dataset containing annotated six scenes with the dominant glass plane and one additional office building with multiple small glass planes captured by RIEGL VZ-400~\citep{RIEGLVZ4002014}
; 3DRN Dataset containing both outdoor urban scenes (two street-view scenarios) captured by RIEGL VZ-2000i~\citep{RIEGLVZ2000i2024}.

Our evaluation consists of two parts: glass plane estimation and virtual point detection.
For glass estimation, we compare our proposed method with the multi-echoes count method~\citep{Yun2021Virtual,Shao2026GRASS} and an intensity-based method~\citep{Shao2023Reflections} using the dominant glass scene data and the multiple small glass scene data.
For virtual point detection, we evaluate the performance of the proposed method in comparison with the existing methods \citep{Yun2021Virtual,Yun2019Cluster,Lee2023Learning,Fang2025Coupled, Shao2026GRASS} in outdoor urban scenes.

\subsection{Results of Glass Plane Estimation}
\label{sec:glass_plane_estimation}

\begin{figure*}
    \centering
    \includegraphics[width=\textwidth]{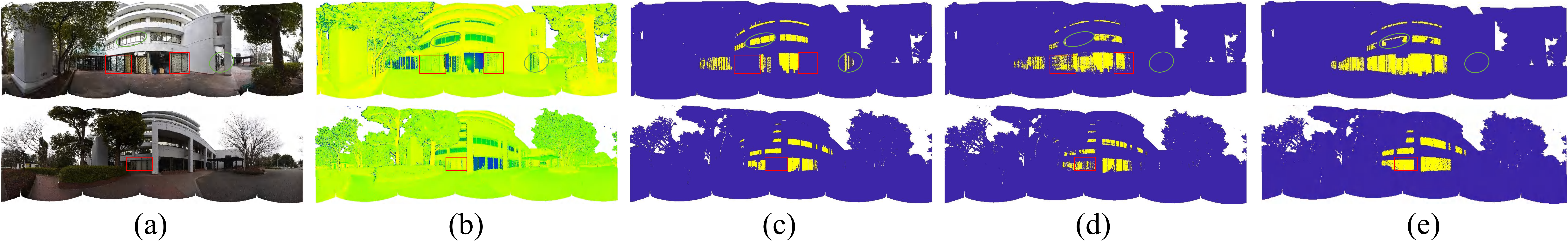}
    \caption{Comparision of the glass region estimation results on TLS data with multiple small glass planes. (a) The panoramic images of input TLS point clouds: The red rectangles are glass objects with curtains drawn behind; green circles indicate glass component with no virtual points. (b) Intensity map; the color from dark blue to bright yellow represents the intensity values, ranging from low to high. (c) Glass regions estimation results based on intensity-based method~\citep{Shao2023Reflections}, where the detected windows are coded in yellow and other points are coded in purple. (d) Glass components estimation based on the multi-echo count method, the red rectangle highlights the sparse pattern~\citep{Yun2018Reflection}. (e) Glass region estimation results by the segment-level multi-echo count method~\citep{Shao2026GRASS}. (f) Glass region estimation results by the proposed algorithm. Results are shown from \textbf{top} to \textbf{bottom} for the scenes: ``Shopping mall'' and ``Office building''.}
    \label{fig:multi_glass_extraction}
\end{figure*}
\begin{figure*}
    \centering
    \includegraphics[width=\textwidth]{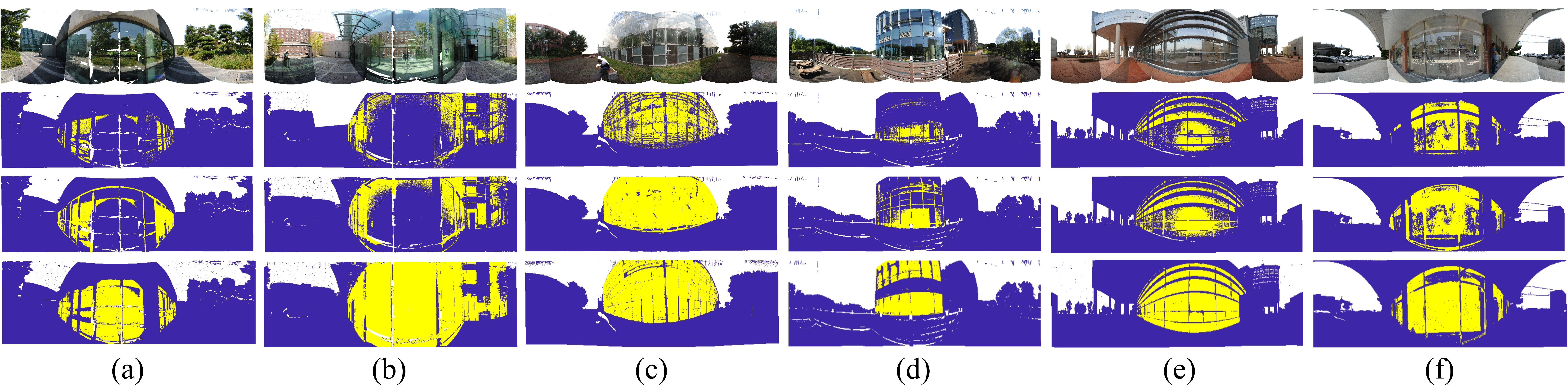}
    \caption{Comparison of the dominant glass region estimation. First row: panoramic images of input point clouds. Second row: multi-return method~\citep{Yun2018Reflection}. Third row: segment-level multi-count method~\citep{Shao2026GRASS}. Bottom row: our proposed method. Scenes from left to right: (a) “Architecture building”, (b) “International hall”, (c) “Botanical garden”, (d) “Terrace”, (e) “Engineering building”, (f) “Gymnasium”.}
    \label{fig:dominant_glass_extraction}
\end{figure*}

In this section, we first evaluate glass plane estimation in scenes containing multiple small glass planes.
Our primary goal is to highlight the limitations of the intensity-based glass filtering method proposed by~\citep{Shao2023Reflections}, particularly its sensitivity to surface reflectance properties. In real-world settings, small windows are often covered by curtains, which effectively exposes the weakness of reflectance-dependent approaches such as~\citep{Shao2023Reflections,Fang2025Coupled}, as they rely on consistent surface reflectivity.

Subsequently, our method focuses on comparing with the multi-return approach for dominant glass plane estimation. The intensity-based method is not included in this comparison. The reason is that, behind large glass surfaces, the points are densely distributed and all exhibit low reflectance due to the energy loss of the laser pulse after transmitting through the glass. Therefore, using intensity alone to detect them becomes inapplicable.

\textbf{Multiple small glass plane scenes:} Fig.~\ref{fig:multi_glass_extraction} shows that the intensity-based method~\citep{Shao2023Reflections} has two limitations: (1) It cannot extract glass objects with curtains drawn behind due to their high intensities returned; as a result, even the points behind these glass objects are all virtual points, but cannot be removed. (2) {It will detect all glass with low intensities, even some of them do not generate reflection-induced virtual points due to scanning geometry (e.g., boundary-of-FOV or no-intersection cases), which will increase the computational load of subsequent processing.} By contrast, the multi-return method~\citep{Yun2018Reflection}, the segment-level multi-count method~\citep{Shao2023Reflections}, and our method can all detect high-intensity glass regions. Nevertheless, our method extracts glass objects that are significantly cleaner and more complete than the other two.

\begin{figure*}
    \centering
    \includegraphics[width=\textwidth]{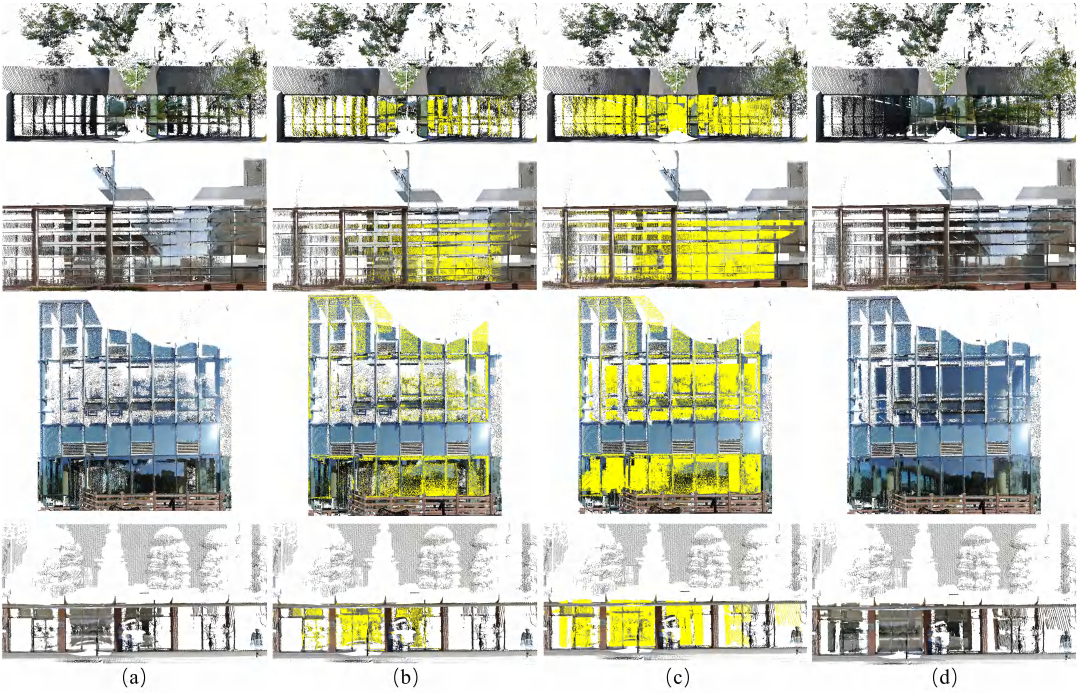}
    \caption{Comparison of glass detection results in several TLS data scenes with severe glass points missing. Detected glass points are shown in yellow in 3D form. It is noted that points close to the glass are not shown to improve the visualization of glass points. (a) Input glass points. (b) Segment-level multi-count results~\citep{Shao2026GRASS}. (c) Our proposed method (d) Target glass points. Results are shown from top to bottom for the scenes: "Architecture building", "Engineering building" , "Terrace", "Gymnasium".}
    \label{fig:glass_detection_results_in_3D}
\end{figure*}

\textbf{Dominant glass plane scene}: As clearly shown in Fig.~\ref{fig:dominant_glass_extraction}, the method of~\citep{Yun2018Reflection}, which relies solely on multi-count information, fails to capture the entire glass areas completely. In such cases, virtual points returned in the missing regions are not considered in the subsequent virtual point removal process. The segment-level multi-count method~\citep{Shao2023Reflections} performs better: it can extract sparse and fragmented glass points to fill in some local missing glass regions. However, there are still numerous holes that cannot be filled due to the absence of a large portion of glass points. In contrast, only our method is capable of completing these large glass holes. This capability can significantly enhance the effectiveness of virtual point filtering. To highlight our method's performance, we also select four scenes with severe glass points missing and show their 3D results in Fig.~\ref{fig:glass_detection_results_in_3D}. The results show that our method significantly fills a large portion of glass voids.

\subsection{Results of Virtual Point Removal}
\label{sec:results_of_virtual_point_removal}
After extracting the glass, we then focus on virtual point removal. After extracting the glass components, virtual point removal is performed. The parameters $\beta_1$ and $\beta_2$ in Eqs.~\eqref{eq:symmetry} and \eqref{eq:similarity}, respectively, together with the local and global support radii $r_1$ and $r_2$ of the RE-LGGS descriptor, are empirically adjusted according to scene complexity. For scenes dominated by man-made structures, we use $\beta_1=2$, $\beta_2=1$, $r_1=0.25$ m, and $r_2=0.5$ m to emphasize fine-scale geometric details. For vegetation-rich environments, larger support radii ($r_1=0.5$ m and $r_2=1.0$ m) together with $\beta_1=\beta_2=0.5$ are adopted to improve robustness against irregular local structures and sparse observations.

\begin{table*}[H]
\setlength{\tabcolsep}{5pt}  
\normalsize
\caption{Quantitative performance comparison in terms of the overall $F_1$ score evaluated on the UNIST dataset (a–e) and an additional multiple small glass building scene (f). Specifically, (a)–(e) correspond to "Architecture building", "Botanical garden", "Engineering building", "Natural science building", and "Terrace", while (f) corresponds to "Office building". The best and second-best values are highlighted in red and blue, respectively.\label{tab:UNIST_results}}
\begin{tabular}{lccccccc}
\toprule
\textbf{Method} & \textbf{(a)} & \textbf{(b)} & \textbf{(c)} & \textbf{(d)} & \textbf{(e)} & \textbf{(f)} & \textbf{Average} \\
\midrule
~\cite{Yun2019Cluster} & 0.615 & 0.585 & 0.729 & 0.880 & 0.520 & 0.709 & 0.673 \\
~\cite{Yun2021Virtual} & 0.694 & 0.822 & 0.627 & 0.777 & 0.379 & -- & 0.659 \\
~\cite{Lee2023Learning} & 0.766 & \textcolor{red}{0.862} & 0.781 & 0.924 & 0.862 & -- & 0.839 \\
~\cite{Shao2026GRASS} & \textcolor{blue}{0.822} & 0.846 & \textcolor{red}{0.923} & \textcolor{red}{0.976} & \textcolor{blue}{0.875} & \textcolor{blue}{0.811} & \textcolor{blue}{0.876} \\
Proposed & \textcolor{red}{0.926} & \textcolor{blue}{0.857} & \textcolor{blue}{0.889} & \textcolor{blue}{0.964} & \textcolor{red}{0.895} & \textcolor{red}{0.816} & \textcolor{red}{0.891} \\
\bottomrule
\end{tabular}

\vspace{4pt}
\noindent\raggedright{\footnotesize{``-'': indicates that the corresponding method is not applicable to the office "building scene" or cannot be faithfully reproduced under the original method assumptions.}}
\end{table*}

\begin{figure*}[H]
\begin{adjustwidth}{-\extralength}{0cm}
    \centering
    \includegraphics[width=0.98\linewidth]{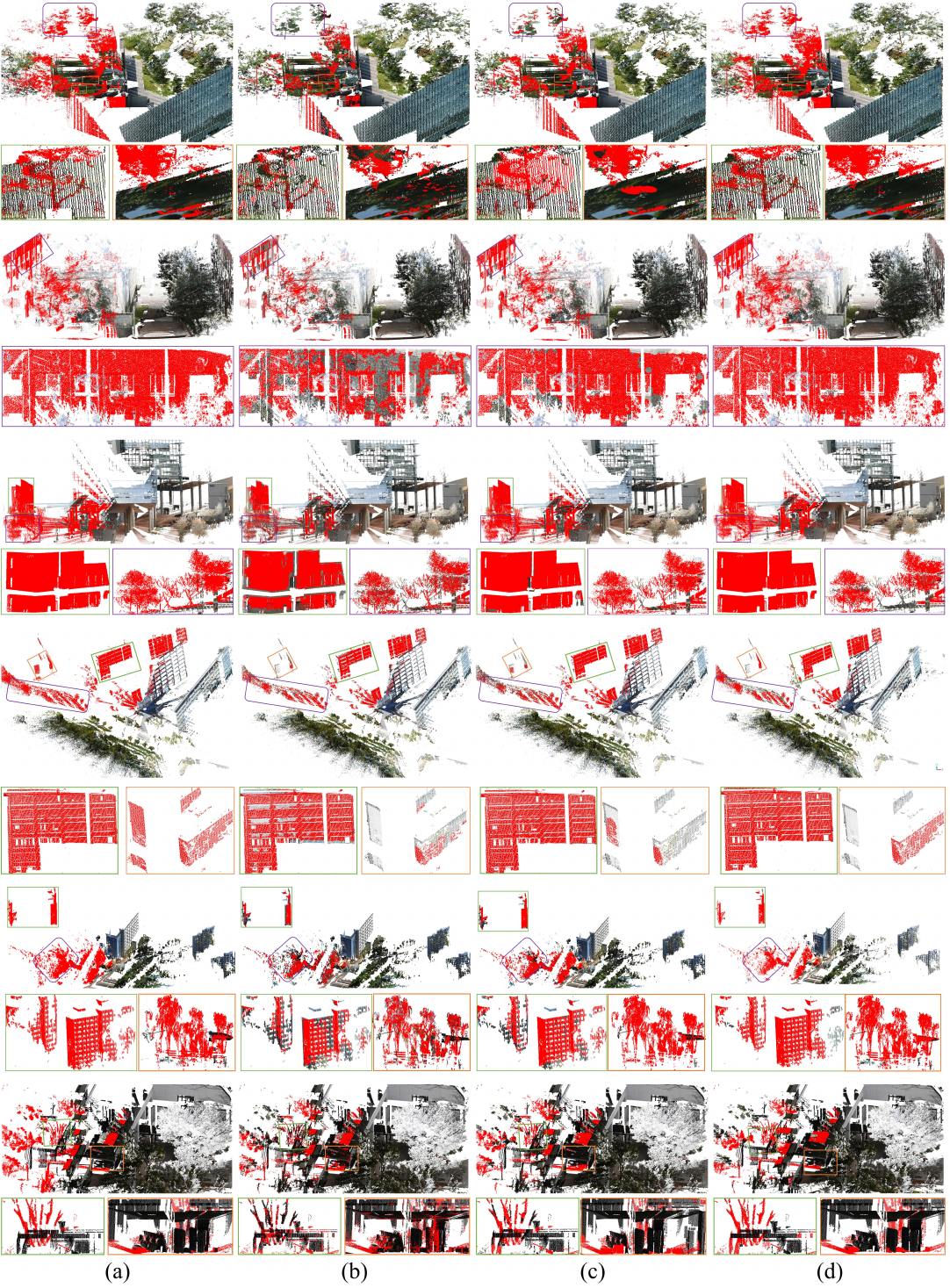}
 \end{adjustwidth}
    \caption{
        Comparison of the virtual point detection results on UNIST building dataset and a multiple-glass dataset.
        (a) Input ground truth points.
        (b) ~\cite{Yun2018Reflection}.
        (c) GRASS~\citep{Shao2026GRASS}.
        (d) Our proposed method. Scenes from top to bottom:
“Architecture building”, “Botanical garden”, “Engineering building”, “Natural science building”,
“Terrace”, “Office building”.
    }
    \label{fig:UNIST_result}
\end{figure*}
\textbf{Results on UNIST dataset:} In Table~\ref{tab:UNIST_results}, we compare the proposed algorithm with other methods based on the $\text{F}_1$ score, given by the following:
\begin{equation} \label{eq:f1_score}
\text{F}_1 = 2 \times \frac{\text{Precision} \times \text{Recall}}{\text{Precision} + \text{Recall}}
\end{equation}
where recall and precision are defined as $\text{Recall} = \text{TP}/(\text{TP} +\text{FN})$ and $\text{Precision} = \text{TP}/(\text{TP} + \text{FP})$, where TP, FP, and FN denote the number of correctly detected virtual points, the number of incorrectly detected virtual points, and number of incorrectly detected valid points, separately. Overall, the proposed method achieves optimal or near‑optimal performance in most cases. As shown in Fig.~\ref{fig:UNIST_result}, our method achieves superior performance on building virtual points, outperforming ~\cite{Yun2018Reflection} and slightly outperforming~\cite{Shao2023Reflections}. Conversely, our method exhibits variable performance in unstructured natural scenarios, particularly within dense vegetation and foliage, as further discussed in sec.~\ref{sec:Discussion}. Due to its segmentation‑based removal strategy, the proposed method also performs well on tree trunk virtual points and avoids misclassifying ground‑near or virtual‑point‑near real indoor objects. 
Despite relying only on lightweight PCA-derived geometric features, the proposed method achieves competitive performance compared with both computationally intensive handcrafted geometric descriptors descriptors and learning-based approaches.

\begin{table*}[H]
\caption{Comparison of different methods on 3DRN datasets. ``Yun'', ``Fang'', and ``Shao'' denote the methods proposed by~\citep{Yun2021Virtual}, \citep{Fang2025Coupled}, and \citep{Shao2026GRASS}, respectively. The best and second-best values are highlighted in red and blue, respectively.}
\label{tab:3DRN_results}
\small
\setlength{\tabcolsep}{2.5pt} 
\begin{adjustwidth}{-\extralength}{0cm}
\centering
\begin{tabularx}{\fulllength}{c*{16}{C}} 
\toprule
& \multicolumn{4}{c}{\textbf{ODR (\%)}} 
& \multicolumn{4}{c}{\textbf{IDR (\%)}} 
& \multicolumn{4}{c}{\textbf{Accuracy (\%)}} 
& \multicolumn{4}{c}{\textbf{SNR (dB)}} \\
\cmidrule(lr){2-5} \cmidrule(lr){6-9}\cmidrule(lr){10-13} \cmidrule(l){14-17}
\raisebox{0.5ex}[0pt]{} & \textbf{Yun} & \textbf{Fang} & \textbf{Shao} & \textbf{Ours}
& \textbf{Yun} & \textbf{Fang} & \textbf{Shao} & \textbf{Ours}
& \textbf{Yun} & \textbf{Fang} & \textbf{Shao} & \textbf{Ours}
& \textbf{Yun} & \textbf{Fang} & \textbf{Shao} & \textbf{Ours}\\
\midrule
Scan 04 & 75.12 & 87.30 & \textcolor{blue}{89.72} & \textcolor{red}{90.57} & \textcolor{red}{99.28} & 98.68 & \textcolor{blue}{98.78} & 95.72 & 91.42 & 95.00 & \textcolor{blue}{95.83} & \textcolor{red}{95.97} & 8.96 & 11.28 & \textcolor{blue}{12.09} & \textcolor{red}{12.41} \\
Scan 05 & 50.24 & 88.53 & \textcolor{blue}{90.52} & \textcolor{red}{94.15} & 97.49 & 97.48 & \textcolor{blue}{97.55} & \textcolor{red}{98.57} & 88.32 & 95.74 & \textcolor{blue}{96.18} & \textcolor{red}{96.56} & 8.39 & 12.77 & \textcolor{blue}{13.25} & \textcolor{red}{13.63} \\
average & 67.65 & 72.71 & \textcolor{blue}{85.07} & \textcolor{red}{92.36} & 87.45 & 93.75 & \textcolor{blue}{94.74} & \textcolor{red}{97.14} & 82.08 & 89.66 & \textcolor{blue}{93.05} & \textcolor{red}{96.27} & 6.98 & 10.03 & \textcolor{blue}{11.14} & \textcolor{red}{13.02} \\
\bottomrule
\end{tabularx}
\end{adjustwidth}
\end{table*}

\begin{figure*}[H]
\begin{adjustwidth}{-\extralength}{0cm}
    \centering
    \includegraphics[width=\linewidth]{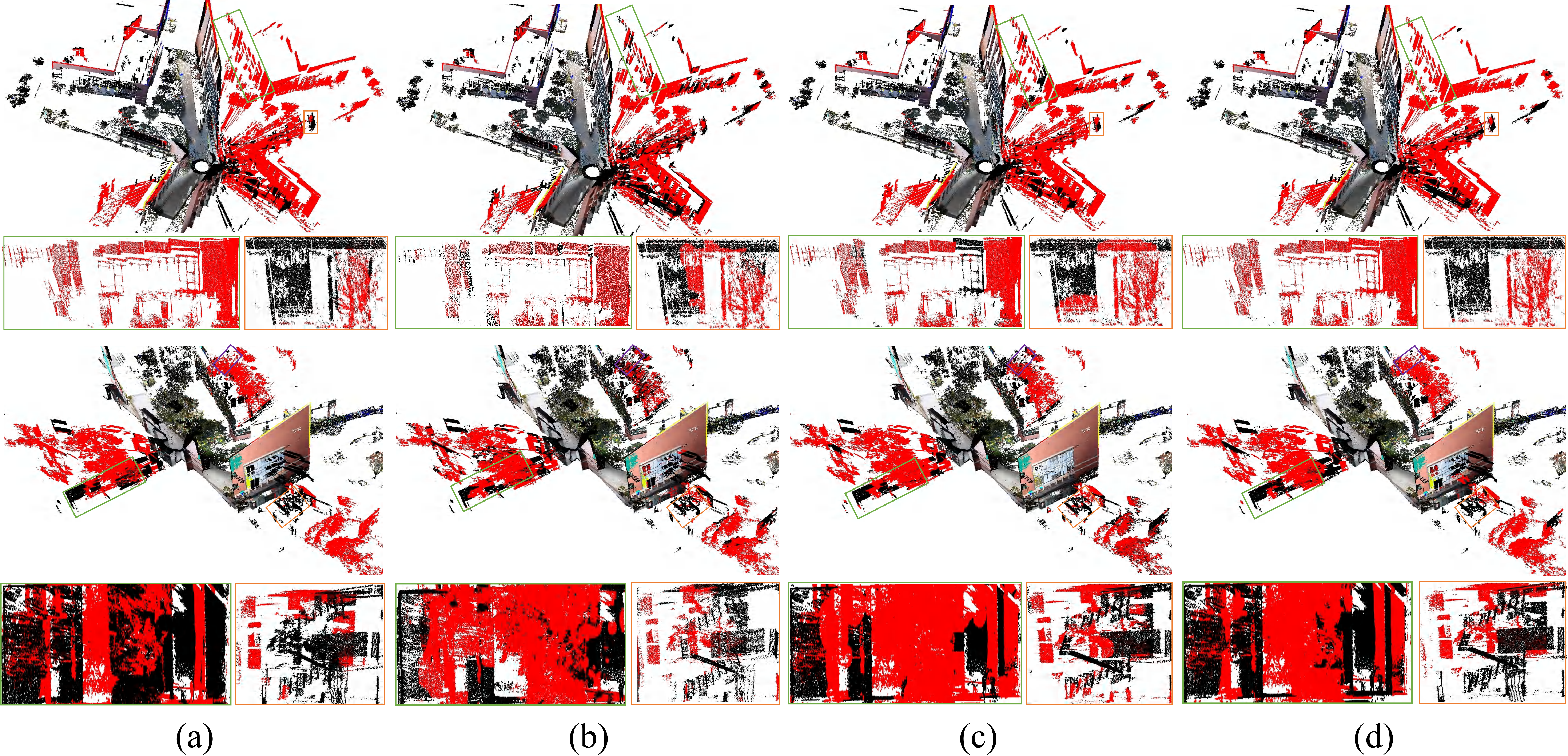}
 \end{adjustwidth}
    \caption{
        Result of proposed method on two "street view" 3DRN dataset. The red points denote virtual points.
        (a) Input ground truth points.
        (b) Our implementation of the method from \cite{Fang2025Coupled} (provided for reference).
        (c) GRASS~\citep{Shao2026GRASS}.
        (d) Our proposed method. Scenes from top to bottom: "Scan 04", "Scan 05".
    }
    \label{fig:3DRN_results}
\end{figure*}

\textbf{Results on 3DRN dataset:} To ensure direct comparability with previously reported results on the 3DRN benchmark, we follow the evaluation protocol adopted by \cite{Fang2025Coupled}, which reports four metrics: outlier detection rate (ODR), inlier detection rate (IDR), accuracy, and signal-to-noise ratio (SNR):
\begin{align}
\text{ODR} &= \frac{\text{TN}}{\text{FP} + \text{TN}} \label{eq:odr} \\
\text{IDR} &= \frac{\text{TP}}{\text{TP} + \text{FN}} \label{eq:idr} \\
\text{Accuracy} &= \frac{\text{TP} + \text{TN}}{\text{TP} + \text{FN} + \text{FP} + \text{TN}} \label{eq:accuracy} \\
\text{SNR} &= 10\cdot\lg\frac{\text{TP} + \text{FN}}{\text{FP} + \text{FN}} \label{eq:snr}
\end{align}
In these metrics, TN represents the number of correctly removed virtual points, while FN describes the number of real-world points incorrectly detected as virtual points. TP refers to the number of preserved real points, and FP is the number of undetected virtual points. In these evaluation metrics, the ODR function corresponds to recall in the $\text{F}_1$ score metric system, measuring the proportion of detected virtual reflection points out of all {virtual points}. The IDR denotes the ratio of preserved real-world points to all ground-truth real points. The accuracy reflects the overall efficacy of the proposed method in eliminating virtual points; SNR quantifies the denoised output quality in decibels: A higher SNR indicates superior signal quality with negligible noise interference. As shown in Table~\ref{tab:3DRN_results}, our method achieves the best performance. From Fig.~\ref{fig:3DRN_results}, it can be observed that in Scan 04, our method successfully identifies virtual points building facades even when they exhibit significant misalignment distortion, and incompleteness. In Scan 05, due to our segmentation‑based virtual point removal strategy, indoor walls partially covered by  vegetation virtual points and stair bottoms close to the ground are not erroneously classified as virtual points.

\begin{figure}
    \centering
    \includegraphics[width=0.99\columnwidth]{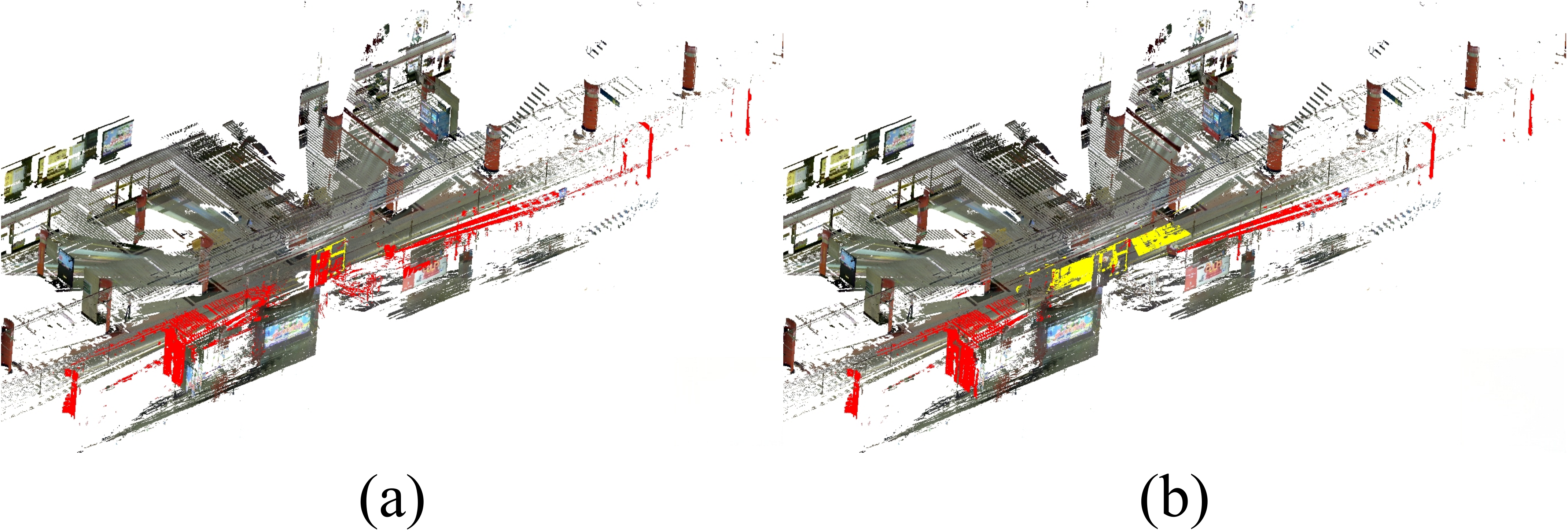}
    \caption{Reflection artifact removal in a subway scene~\citep{DONG2020327}, where virtual points appear on the tracks. (a) Manually annotated reference data; glass points are shown in yellow and virtual points in red. (b) Virtual point removal results produced by the proposed method.}
    \label{fig:subway_results}
\end{figure}
Finally, we also apply the proposed method on a subway dataset, as shown in Fig.~\ref{fig:subway_results}. Due to measurement voids, the raw point cloud contains almost no return points from the glass surfaces (Fig.~\ref{fig:subway_results}(a)). This scarcity makes it exceptionally challenging to detect glass using geometry information alone, let alone eliminate the resulting reflection artifacts. In contrast, by combining image and LiDAR information with geometric constraints, the missing glass surfaces can be recovered, and most reflection-induced virtual points can be identified and removed (Fig.~\ref{fig:subway_results}(b)).
\section{Discussion}
\label{sec:Discussion}

\noindent $\bullet$ \textbf{\textit{Why employ a vision foundation model for glass detection?}} 

The adoption of a vision foundation model (i.e., Nano Banana) is not intended to demonstrate the superiority of a particular foundation model, but rather to exploit the emerging capability of vision foundation models to bridge the observation gap between image semantics and TLS geometry.
In glass-rich TLS scenes, the major challenge is not geometric reconstruction itself, but the reliable localization of glass regions under severe measurement incompleteness. Traditional geometry-based methods often fail because glass surfaces produce sparse, fragmented, or even missing returns. By jointly reasoning over panoramic RGB imagery, intensity maps, and multi-count maps, Nano Banana provides a high-completeness semantic glass prior that is difficult to obtain from point clouds alone.
Therefore, the contribution of this work lies less in the foundation model itself and more in demonstrating how vision foundation models can be integrated into geometric surveying workflows for robust glass-region estimation.

\noindent $\bullet$ \textbf{\textit{Why does RE-LGGS can describe imperfect reflections?}}

Existing reflection artifact removal methods typically assume ideal reflection symmetry, where virtual points remain geometrically consistent with their real counterparts. In practice, however, glass reflections are often affected by geometric incompleteness, spatial distortions, and many-to-one correspondence ambiguities. To address these challenges, the proposed RE-LGGS descriptor emphasizes structural robustness under imperfect reflection observations. Specifically, a dual-scale collaborative mechanism is adopted, where small-radius neighborhoods extracted from complete real points preserve fine-scale geometric characteristics, while large-radius neighborhoods extracted from searched real points capture more stable structural tendencies. The descriptor further characterizes local geometry using PCA-derived linearity, planarity, and curvature, which remain relatively stable under sparse and degraded observations. In addition, an adaptive orientation consistency constraint is introduced to distinguish structures with similar geometric statistics but different spatial organizations. By jointly integrating multi-scale geometric information and orientation consistency, RE-LGGS provides a robust and physically interpretable similarity measure for virtual point identification. The experimental results demonstrate that the descriptor remains effective even when ideal reflection symmetry is severely violated, highlighting its suitability for real-world TLS reflection artifacts rather than idealized mirror-reflection assumptions.

\noindent $\bullet$ \textbf{\textit{Why does our method perform unstable in dense vegetation scenes?}}

Although the proposed framework achieves strong performance in structured man-made environments, minor performance degradation can be observed in dense vegetation regions. This limitation mainly originates from the intrinsic structural inconsistency between virtual and real vegetation observations. Unlike architectural objects with stable geometric primitives, vegetation exhibits highly irregular and heterogeneous spatial distributions. Due to laser energy attenuation, partial reflections, and incomplete sampling, virtual vegetation points often preserve only sparse outer-surface structures, whereas their corresponding real observations remain volumetrically distributed. Consequently, the two point sets may exhibit substantially different covariance statistics and PCA-derived shape attributes. For example, a virtual vegetation patch may appear locally planar, while its real counterpart exhibits a highly scattered or isotropic distribution. Such structural discrepancies reduce the consistency of the RE-LGGS descriptor and occasionally lead to ambiguous similarity measurements. Nevertheless, the proposed method still maintains competitive performance in these challenging scenarios while relying only on lightweight geometric representations and without requiring any training data.

\textbf{Future work:} (1) \textit{Non-planar glass modeling}: Future research will extend the current plane-based detection framework to handle curved, irregular, or discontinuous glass surfaces (e.g., cylindrical storefronts and complex architectural facades), thereby broadening the applicability of the proposed method. (2) \textit{Multiple-reflection reasoning}: Future research will investigate more challenging scenarios involving multiple reflective glasses, where repeated reflections between glass surfaces generate complex virtual structures. Such situations frequently occur in large indoor spaces containing multiple glass partitions, commercial streets with opposing shopfronts, and large glass curtain-wall systems. Developing robust mechanisms for modeling and disentangling these higher-order reflections remains an important research direction.(3) \textit{Generalization across sensing platforms}: Future algorithms should possess stronger adaptability to open-world environments and diverse acquisition platforms. Beyond static TLS systems, reflection artifact removal should be extended to mobile mapping platforms, including vehicle-mounted, backpack, and UAV-based LiDAR systems. Achieving a favorable balance between computational efficiency and geometric accuracy will be essential for processing large-scale point clouds while maintaining the centimeter-level precision required by digital twin applications. (4) \textit{Foundation-model adaptation and automation}: Although Nano Banana is adopted in this work, the proposed framework is not restricted to a specific foundation model. Future research will investigate the integration of emerging open-source and domain-adapted vision foundation models, as well as automated prompt generation strategies, to further improve robustness and reproducibility in large-scale surveying applications.

\section{Conclusion}
\label{sec:conclusion}

In this paper, we presented a unified framework for glass-induced reflection artifact removal in TLS point clouds. The proposed framework addresses two fundamental challenges in glass-rich TLS environments: incomplete glass observations and imperfect reflection-induced virtual points.
To recover missing glass surfaces, we integrate a vision foundation model with geometric refinement and completion strategies, enabling reliable glass-region estimation even under severe measurement voids. To identify reflection artifacts, we introduce a Reflection-aware Local-Global Geometric Similarity (RE-LGGS) descriptor, which combines dual-scale geometric statistics and adaptive orientation consistency to achieve robust virtual point detection under geometric incompleteness, spatial distortions, and many-to-one correspondence effects caused by real-world reflections.
Extensive experiments on multiple public datasets demonstrate that the proposed framework consistently outperforms existing state-of-the-art methods in both glass estimation and reflection artifact removal tasks. The results further indicate that combining foundation-model semantic perception with physically interpretable geometric reasoning provides a practical and effective solution for improving TLS data quality in complex urban environments. We believe that this paradigm offers a promising direction for integrating vision foundation models into future surveying and mapping workflows.

\section{Acknowledge}
\label{sec:funding}
This work is supported by the National Key R\&D Program of China (No.2022ZD0119000), 
Hunan Provincial Key R\&D Program of China (No.2024JK2020 and 2024JK2021), 
Hunan Provincial Natural Science Foundation of China (No.2024JJ10027), 
Young Talents of Huxiang (No.Z202433000575), 
and Changsha Science Fund for Distinguished Young Scholars (kq2306002).








\printcredits

\bibliographystyle{cas-model2-names}

\bibliography{cas-refs}



\end{document}